\documentclass[journal, 10pt]{IEEEtran}

\usepackage{cite}
\usepackage{amsmath,amssymb,amsthm}
\usepackage{graphicx}
\usepackage{moreverb,url}
\usepackage[colorlinks,bookmarksopen,bookmarksnumbered,citecolor=red,urlcolor=red]{hyperref}
\usepackage[linesnumbered, ruled]{algorithm2e}
\usepackage{subfig}
\usepackage{multirow, booktabs}
\usepackage{cuted}

\theoremstyle{definition}

\theoremstyle{remark}

\newcommand{\ie}{\textit{i}.\textit{e}., }
\newcommand{\etc}{\textit{etc}}
\newcommand{\eg}{\textit{e}.\textit{g}., }
\newcommand{\IR}{\mathbb{R}}
\newcommand{\II}{\mathbb{I}}
\newcommand{\oomega}{\boldsymbol{\omega}}
\newcommand{\xxi}{\boldsymbol{\xi}}
\newcommand{\SE}{\text{SE}}
\newcommand{\SO}{\text{SO}}
\newcommand{\PCG}{\text{PCG}}

\newcommand{\xx}{\boldsymbol{x}}
\newcommand{\yy}{\boldsymbol{y}}

\newtheorem{theorem}{Theorem}[section]

\begin{document}

\title{PRIMP: PRobabilistically-Informed Motion Primitives for Efficient Affordance Learning from Demonstration}
\author{Sipu Ruan, Weixiao Liu, Xiaoli Wang, Xin Meng and Gregory S. Chirikjian*
\thanks{Sipu Ruan (\texttt{ruansp@nus.edu.sg}), Xiaoli Wang, Xin Meng and Gregory S. Chirikjian (\texttt{mpegre@nus.edu.sg}) are with Department of Mechanical Engineering, National University of Singapore, Singapore.}
\thanks{Weixiao Liu is with Department of Mechanical Engineering, Johns Hopkins University, Baltimore, MD, USA.}
\thanks{* Address all correspondence to this author.}}

\maketitle

\begin{abstract}
This paper proposes a learning-from-demonstration method using probability densities on the workspaces of robot manipulators. The method, named ``PRobabilistically-Informed Motion Primitives (PRIMP)'', learns the probability distribution of the end effector trajectories in the 6D workspace that includes both positions and orientations. It is able to adapt to new situations such as novel via poses with uncertainty and a change of viewing frame. The method itself is robot-agnostic, in which the learned distribution can be transferred to another robot with the adaptation to its workspace density. The learned trajectory distribution is then used to guide an optimization-based motion planning algorithm to further help the robot avoid novel obstacles that are unseen during the demonstration process. The proposed methods are evaluated by several sets of benchmark experiments. PRIMP runs more than 5 times faster while generalizing trajectories more than twice as close to both the demonstrations and novel desired poses. It is then combined with our robot imagination method that learns object affordances, illustrating the applicability of PRIMP to learn tool use through physical experiments.
\end{abstract}


\section{Introduction} \label{sec:introduction}
\begin{figure}
\centering
\includegraphics[scale=0.55, trim = 250 100 275 100, clip]{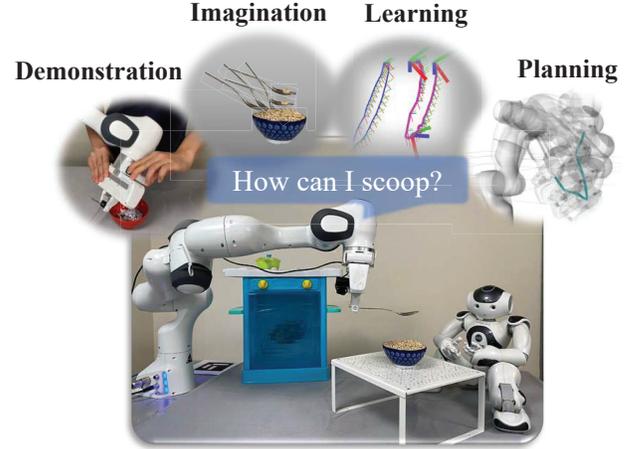}
\caption{Illustration for the general idea of this work. The robot arm is asked to use a spoon to scoop from a bowl in a household environment. With the help of human demonstrations, imagination of object affordance, learning skills from the demonstrations and motion planning, the robot is able to fulfill the task in a novel scene with unseen obstacles.}
\label{fig:introduction:cover}
\end{figure}

For a robot to be truly intelligent, it needs the ability to learn from prior knowledge while adapting to unseen scenarios. The prior knowledge can be some previous feasible trajectories that are hard-coded, like in a structured factory environment. Prior knowledge can also be from human-demonstrated motions, which are difficult to pre-program but are ubiquitous in household environments, like scooping powder (as in Fig. \ref{fig:introduction:cover}). The latter case is much more challenging and related to a popular field in robot learning, namely \textit{Learning-from-Demonstration (LfD)} \cite{ravichandar2020recent} or \textit{Programming-by-Demonstration (PbD)} \cite{billard2008robot}. Many works on LfD encode demonstrations as trajectories in Euclidean space \cite{jin2022learning,zhu2022learning}. For example, the angles and velocities of all joints are considered as a multi-dimensional state vector; or only the positions of the end effector are modeled. However, the models in the full workspace (including both position and orientation) of a robot manipulator have not been considered until recently \cite{zeestraten2017approach,calinon2020gaussians,huang2020toward,ti2023geometric,auddy2023continual}. Our work focuses on the robot workspace and proposes a novel method using probability densities on Lie groups, denoted as \textit{PRobabilistically-Informed Motion Primitives (PRIMP)}. The mathematical model is inspired by a concept initially introduced in our group more than 25 years ago and the concept of loop entropy \cite{chirikjian2000engineering,chirikjian2011modeling} and more recent work on inverse reachability mapping \cite{vahrenkamp2013robot,jauhri2022robot}. It is further extended here into LfD with via-point conditioning as compared to only subjecting to end constraints. The learned knowledge is robot-agnostic but can be adapted to the workspace of a specific robot. It is also able to deal with extrapolation cases and model with a few or even a single demonstration.

Demonstrations are typically conducted in a scene with only the robot and the object it interacts with. However, when there are some novel obstacles unseen during the demonstrations, only using LfD methods is not enough to generate feasible trajectories. Here is where the motion planning process fits in. It is able to guarantee safe and optimal motions for a task, which is formulated as motion constraints. For most motion planners, either sampling-based or optimization-based, constraints are added as a function with specific expressions. For example, to require holding a cup that is filled with water upright, one usually needs to define the vector that is perpendicular to the cup opening to be parallel with the global $z$-axis. Such an expression is, however, not trivial to define for more complex tasks, \eg scooping powders and writing letters. For such tasks, LfD can play the role of guiding the motion planners via a trajectory-wise cost function.

Guided motion planning, which combines LfD and motion planning, has become popular in the recent decade\cite{koert2016demonstration,ye2017guided,magyar2019guided,dobivs2021cartesian}. The goal is to make the motion collision-free while keeping the critical features from the learned trajectory as many as possible. In this work, an optimization-based planner, \textit{Stochastic Trajectory Optimization for Motion Planning (STOMP)} \cite{kalakrishnan2011stomp}, is applied as the base framework. The learned trajectory distribution via PRIMP is used as a reference. It is firstly treated as the initial condition for the planner. Then, a novel cost function with respect to this reference distribution is proposed to guide the planning process. Instead of joint space, the cost function is based on the workspace of the robot end effector. Therefore, the planner is named as \textit{Workspace-STOMP}.

Apart from novel obstacles, the object that the robot interacts with might also be unseen. For example, the robot has been demonstrated how to pour powders from a cup into a bowl. In this case, the cup is treated as a tool for pouring and the bowl has the affordance of containing. In a new scenario, the tool for pouring might be changed into a spoon and the object becomes a vase, which has the same affordance of containing. The learned trajectory distribution should be able to adapt to this new situation with the same set of demonstrations, even when the appearances and categories of the tool and object are totally different. In order to fulfill similar tasks intelligently, the understanding of object functionality and affordance is a key aspect \cite{ho2022general}. Recent works proposed methods to learn object affordance, like pouring, via physics-based simulation \cite{wu2020can,vosylius2022start}. Our work applies this idea to learn the key points of a task, \ie the pouring pose, which is used as a new goal or via pose for re-production. A new task that learns the affordance of scooping is proposed and implemented in the simulation environment. The affordance learning of an object using simulation is then combined into a robotic system with PRIMP and Workspace-STOMP through physical experiments.

The contributions of our work are listed as follows.
\begin{itemize}
\item A learning-from-demonstration method, PRIMP, is proposed to generate reference workspace trajectory distribution for basic motion primitives;

\item By proposing a novel cost function related to the reference distribution, Workspace-STOMP is proposed to keep the shape of the trajectory similar while maintaining the feasibility of the plan.

\item A novel robotic system that combines LfD, motion planning, and affordance-based simulation is proposed and physically demonstrated in a robot manipulator platform.
\end{itemize}

The rest of this paper is organized as follows. Section \ref{sec:literature} reviews related literature on LfD and guided motion planning. Section \ref{sec:PRIMP} introduces the proposed PRIMP method. Section \ref{sec:wstomp} proposes the Workspace-STOMP motion planning algorithm. Then, Sec. \ref{sec:benchmark} evaluates PRIMP and Workspace-STOMP through benchmarks with existing probabilistic methods for solving LfD and planning problems. Section \ref{sec:physical-experiments} proposes a robotic system that combines PRIMP and Workspace-STOMP with affordance-based simulation and demonstrates via several physical experiments. The proposed method is discussed in Sec. \ref{sec:discussion}. Finally, this paper is concluded in Sec. \ref{sec:conclusion}.

\section{Literature Review} \label{sec:literature}
Learning-from-Demonstration (LfD) is a fast-growing field in robot learning. It aims for robots to imitate human experts while adapting to new situations. A growing amount of effort shows that the combination of LfD and motion planning is able to adapt to a broader field of scenarios where unseen obstacles appear. The following reviews related work on LfD methods and guided motion planning algorithms.

\subsection{Learning from Demonstration}
Over the past decade, many LfD methods have been proposed to encode trajectory models and generalize them to new situations, most of which stay in Euclidean space. Dynamical movement primitives (DMP) \cite{ijspeert2013dynamical} learns a damped spring model to obtain a mapping between acceleration and position-speed pair. It is able to represent and encode goal-directed or periodic trajectories given any start and goal points, but cannot generalize to pass via points. By combining the Gaussian process, DMP exploits uncertainty information during reproduction \cite{fanger2016gaussian}. The parameters, \ie the global time scale, weights of the forcing term, \etc, can be learned using reinforcement learning techniques \cite{li2017reinforcement}. To further solve for the violation of joint limits caused by exploration noise, a constrained DMP (CDMP) is proposed \cite{duan2018constrained}. 

Another class of LfD methods is based on probabilistic model, which is the category that our proposed method falls in. Task-parameterized Gaussian mixture model (TP-GMM) \cite{calinon2014task,calinon2016tutorial} encodes the demonstrations using virtual spring-damper systems in multiple reference frames. It learns the parameters in an expectation-maximization fashion and is able to reproduce in new situations, such as adaptation to via points. Probabilistic movement primitives (ProMP) \cite{paraschos2018using} learns the distribution given a set of demonstrated trajectories and uses conditioning to generalize to novel situations such as via-points and temporal variances. The learning phase is based on a hierarchical Bayesian model, which requires the definition of basis such as radial and Fourier basis functions. Several variants have been proposed to estimate parameters using fewer training instances \cite{gomez2020adaptation}, unify adaptation of the trajectory using constrained optimization \cite{frank2021constrained}, adapt to external contextual information such as object mass using active learning technique \cite{kulak2021active}, \etc. However, when the data dimension is large, ProMP might require more basis functions to well-encode the demonstrations. To deal with the extrapolation issue, Via-point movement primitives (VMP) \cite{zhou2019learning} combines ProMP with ideas from DMP by splitting the trajectory model into elementary trajectories and shape modulation terms. Kernelized movement primitives (KMP) \cite{huang2019kernelized} uses the kernel trick to deal with high dimensional input data without defining basis functions and can also cope with extrapolation issues. In our work, ProMP is selected in the benchmark studies to only handle the data in Euclidean space, \ie the position information of trajectories.

Recently, more investigations have been conducted on encoding trajectories and generalizing to new situations within a manifold, \ie to precisely describe the orientation of the end effector \cite{calinon2020gaussians}. DMP is extended to encode orientation information, which uses quaternions to provide singularity-free representation \cite{ude2014orientation}. Several probabilistic methods on Riemannian manifolds are also proposed. An extension for the Gaussian mixture model (GMM) with inference using Gaussian mixture regression (GMR) or TP-GMM has been proposed to work on Riemannian manifolds \cite{zeestraten2017approach}. The operations such as Gaussian product and conditioning are defined using Riemannian statistics and exponential mapping. It also has applications for bimanual manipulation \cite{silverio2015learning} and grasping \cite{ti2023geometric} tasks. Operations in ProMP can also work on Riemannian manifold, including trajectory modulation, blending, task parameterization, \etc \cite{rozo2022orientation}. It uses geodesic regression to estimate the weight distribution, which acts as a generalized technique for linear regression in Euclidean space. Probabilistic learning of quaternions is applied in the KMP framework for modeling orientation and angular velocity \cite{huang2020toward}. The trajectory after adaptation is then optimized with angular and acceleration constraints. The Orientation-KMP is used in the benchmark studies. Our proposed LfD method differs in that no basis function or kernel is required. Also, the distribution is defined by the relative poses between adjacent time steps compared to the absolute poses in the trajectory.

\subsection{Guided motion planning}
Guided motion planning can be viewed as a special type of constrained motion planning. The constraints are mostly treated as measures of how close a trajectory is to the guiding reference. A demonstration-guided motion planning (DGMP) framework has been proposed to combine LfD and a sampling-based planner \cite{ye2017guided}. It encodes the reference in joint space with a probabilistic model and records surrounding landmarks for correspondence matching among different scenarios. The reference path is extended around new obstacles by sampling configurations from the probabilistic model and formulating a local graph to search for a new safe and optimal path. A Combined Learning from demonstration And Motion Planning (CLAMP) carries out probabilistic inference to search optimal trajectories for a certain skill that are feasible in novel scenes \cite{rana2018towards}. The trajectory prior is generated by the Gaussian process, which is the solution to a linear time-varying stochastic differential equation. To reproduce trajectories, the algorithm uses maximum a posterior (MAP) inference. The method is extended to conduct demonstrations in a cluttered environment with weighted skill learning \cite{rana2018learning}. Defining cost functions using learned trajectory from the LfD method is popular in guiding optimization-based motion planners. For example, the cost functional of CHOMP \cite{zucker2013chomp} is proposed to measure deviations from the probability distribution of the demonstrations \cite{osa2017guiding}. Alternatively, results from ProMP have been used as initial conditions for CHOMP \cite{shyam2019improving}. For a partially observable environment, the inverse reinforcement learning technique is utilized to derive motion level guidance, the reward of which is used to guide the TrajOpt algorithm \cite{quintero2022human}. STOMP is another popular planner that uses rollouts sampled from probabilistic distribution to avoid gradient computations. Cost functions based on DMP working on joint space (namely GSTOMP) \cite{magyar2019guided} and Cartesian constraints that deals with both position and orientation of end effector (\ie Cartesian-STOMP) \cite{dobivs2021cartesian} have been proposed. This work uses Cartesian-STOMP for benchmarks, but the planning part of our method differs significantly. Instead of using only one trajectory as guidance, ours utilizes the information of the probabilistic distribution of the trajectories. 


\section{PRobabilistically-Informed Motion Primitives} \label{sec:PRIMP}
This section introduces the proposed LfD method, named \textit{PRobabilistically-Informed Motion Primitives (PRIMP)}. In general, this is a probabilistic method that encodes the trajectory that involves both position and orientation of the robot end effector. The trajectory is represented discretely by a user-selected number of time steps (\ie $N_{\rm step}$). Each pose is modeled as an element in Lie group $G$. Therefore, the full state is considered in a product space $G \times ... \times G$, resulting in a state vector of dimension $6 N_{\rm step}$. The widely-recognized group in robotics, \ie $\SE(3)$, is used to derive the proposed method. And the extension to \textit{Pose Change Group, $\PCG(3)$} \cite{chirikjian2018pose}, is also introduced. Without loss of generality and special mentions, $\SE(3)$ is used in the following introductions of the method (some basic operations in $\SE(3)$ and $\PCG(3)$ extensively used in this work are reviewed in Appendices \ref{appendix:se} and \ref{appendix:pcg}, respectively).

Consider a set of demonstrated trajectories $\left\{ g^{(k)}_0, g^{(k)}_1,...,.g^{(k)}_n \right\}$, where $g^{(k)}_i \in \SE(3)$ is the $i^{\rm th}$ step in the $k^{\rm th}$ demonstration. The goal is to compute a probability distribution of the given demonstrations as a reference to guide the future executions of the robot for a similar task. Also, if the robot is asked to pass through several via poses $g^*_i$ with uncertainties described by covariance matrix $\Sigma^*_i$, the learned reference density should be adapted to this new situation. The following derivations assume that $i^{\rm th}$ step is only affected by its neighboring $(i-1)^{\rm th}$ and $(i+1)^{\rm th}$ steps.

\subsection{General framework of PRIMP}
Firstly, all the demonstrated trajectories are temporally aligned into the same time scale, via \textit{Globally-Optimal Reparameterization Algorithm (GORA)} \cite{mitchel2018signal} (Sec. \ref{sec:PRIMP:gora}). After alignment, the probability distribution of the $(i+1)^{\rm th}$ pose with respect to the $i^{\rm th}$ pose is approximated using a Lie-theoretic method (Sec. \ref{sec:PRIMP:relative_distribution}). The computed initial mean and covariance is then encoded into the joint distribution of the whole trajectory (Sec. \ref{sec:PRIMP:joint_distribution}). Several types of adaptations to novel scenarios are then introduced (Sec. \ref{sec:PRIMP:adaptation}): 
\begin{itemize}
\item when an intermediate pose is different from the given mean and has some uncertainties, a posterior distribution is computed to adapt to this new situation (Sec. \ref{sec:PRIMP:adaption:via_point});
\item when there is a change of viewing frame, the encoded distribution can be adapted in the sense of equivariance (Sec. \ref{sec:PRIMP:adaptation:change_view});
\item when another robot is operated for the same task, the learned distribution can be further conditioned to the high density region of the new workspace (Sec. \ref{sec:PRIMP:adaptation:wd}).
\end{itemize}
All the above computations can be extended to $\PCG(3)$, which is discussed in Sec. \ref{sec:PRIMP:pcg}.

\subsection{Temporal Alignment for Demonstrations using Globally Optimal Reparameterization Algorithm (GORA)} \label{sec:PRIMP:gora}
Different demonstrations might have different speed of execution, even when the trajectories have the same shape \cite{jin2022learning}. This misalignment in the temporal axis provides difficulty to the probabilistic modelling. Therefore, a temporal reparameterization of all the demonstrated trajectories is needed. Existing work often used DTW method to temporally align different demonstrations \cite{vakanski2012trajectory,zhu2022learning}. One has to choose one base trajectory and warp others into it, in which different choices of the base trajectory might give different alignment results. Alternatively, in our work, variational calculus technique is applied to align multiple $\SE(3)$ trajectories with global optimality in temporal axis, which is named as Globally Optimal Reparameterization Algorithm (GORA) \cite{mitchel2018signal}.

\subsubsection{Problem formulation}
Suppose an $\SE(3)$ sequence $g(\tau)$ is parameterized by $\tau \in [0,1]$. Here $\tau(t): [0,1] \rightarrow [0,1]$ is a one-dimensional monotonic function that parameterizes time. The total variation of the whole sequence can be computed as the sum of the squared derivative with respect to time $t$, \ie
\begin{equation}
J = \int_{0}^{1} f(\tau, \dot{\tau}) \,dt 
\doteq \int_{0}^{1} \mathfrak{g}(\tau) \, \dot{\tau}^2 \,dt \,,
\label{eq:PRIMP:gora:functional}
\end{equation}
where $\dot{\tau} = d \tau / d t$. The expression of $\mathfrak{g}(\tau)$ in the integrand is defined based on the body velocity of an $\SE(3)$ trajectory, \ie
\begin{equation}
\mathfrak{g}(\tau) \doteq \left\| g^{-1} \frac{\partial g}{\partial \tau} \right\|_W^{2} \,,
\label{eq:PRIMP:gora:integrand}
\end{equation}
where $\left\| \cdot \right\|_W$ denotes the weighted Frobenius norm defined such that for any $\mathcal{A} \in \mathbb{R}^{4 \times 4}$, $\left\| \mathcal{A} \right\|_W = \sqrt{\textrm{tr}\left(\mathcal{A}^T W \mathcal{A}\right)}$, for some symmetric matrix 
$$
W \doteq \left( \begin{matrix} \frac{1}{2}\textrm{tr}(I) \II_3 - I & {\bf 0} \\ {\bf 0}^T & 1 
\end{matrix} \right) \in \mathbb{R}^{4 \times 4} \,,
$$ 
where $I$ is the $3 \times 3$ diagonal inertia tensor corresponding to a solid sphere of unit mass and $\II_3$ is the $3 \times 3$ identity matrix. The structure of the function $f(\tau, \dot{\tau})$ is specifically chosen since the global optimality of the solution can be obtained.

\subsubsection{Globally optimal solution and GORA}
To extremize Eq. \eqref{eq:PRIMP:gora:functional}, the Euler-Lagrange equation is used, which has the form
\begin{equation}
\frac{\partial f}{\partial \tau} - \frac{d}{dt} \left(\frac{\partial f}{\partial \dot{\tau}} \right) = 0 \,.
\label{eq:PRIMP:gora:eulerlag}
\end{equation}
With a special structure of Eq. \eqref{eq:PRIMP:gora:integrand}, the solution of Eq. \eqref{eq:PRIMP:gora:eulerlag} can be proved to be globally minimal, resulting in the following theorem.

\begin{theorem}
The Euler-Lagrange Equation \eqref{eq:PRIMP:gora:eulerlag} globally minimizes Eq. \eqref{eq:PRIMP:gora:functional} if: 
\begin{itemize}
\item the integrand is of the form as Eq. \eqref{eq:PRIMP:gora:integrand};
\item $\tau: [0,1] \, \rightarrow \, [0,1]$ is monotonically increasing with boundary conditions $\tau(0) = 0$ and $\tau(1) =1$. 
\end{itemize}
The minimizer is $\tau^*(t) = F^{-1}(t)$, where $F^{-1}(\cdot)$ is the inverse function of
\begin{equation}
F(\tau^*) = \frac{\int_0^{\tau^*} \mathfrak{g}^{\frac{1}{2}}(\sigma) \, d\sigma}{\int_0^1 \mathfrak{g}^{\frac{1}{2}}(\sigma) \, d\sigma} = t \,.
\label{eq:PRIMP:gora:thm_result}
\end{equation}
\label{thm:PRIMP:gora}
\end{theorem}


The proof of Theorem \ref{thm:PRIMP:gora} can be referred to \cite{mitchel2018signal} and \cite{chirikjian2023bootstrapping}. Algorithm \ref{algo:PRIMP:gora} summarizes the workflow of GORA for $\SE(3)$ sequences. Line \ref{algo:PRIMP:gora:init} initializes the temporal parameter $\tau$ and the original time scale $t$ uniformly within $[0,1]$. The numbers of time steps of $t$ and $\tau$ equal the length of the input sequence and a user-defined number ($N_{\rm step}$) for the reparameterized sequence, respectively. Line \ref{algo:PRIMP:gora:g_tau} computes $\mathfrak{g}(\tau)$ using Eq. \eqref{eq:PRIMP:gora:integrand}. Both Lines \ref{algo:PRIMP:gora:F_tau} and \ref{algo:PRIMP:gora:F_tau_normalized} are the core computations of the algorithm, \ie Eq. \eqref{eq:PRIMP:gora:thm_result}. Here, the trapezoidal rule is applied for numerical integration because of its simplicity. $F(\tau)$ is then normalized by dividing the last element, \ie $F(1)$. Line \ref{algo:PRIMP:gora:tau_opt} obtains the global minimizer $\tau^*(t)$ by inverting $F(\tau^*)$, which is numerically conducted by interpolating $t$ with respect to $F(\tau^*)$. Finally, the input data sequence is reparameterized by $\tau^*$ by numerical interpolation in $\SE(3)$ in Line \ref{algo:PRIMP:gora:x_opt}.
\begin{algorithm}[t]
\SetAlgoLined
\SetKwInOut{Param}{Parameter}
\SetKwInOut{In}{Inputs}
\SetKwInOut{Out}{Outputs}

\In{$g(\tau)$: Input $\SE(3)$ sequence}
\Param{$N_{\rm step}$: Number of time steps}
\Out{$g(\tau^*)$: Optimal reparameterized sequence \\
$\tau^*(t)$: Optimal temporal parameter}

Initialize $\tau$ and $t$ as uniform sampling within $[0,1]$\; \label{algo:PRIMP:gora:init}
Compute $\mathfrak{g}(\tau)$ from Eq. \eqref{eq:PRIMP:gora:integrand}\; \label{algo:PRIMP:gora:g_tau}
\For{time step $i$}{ 
    $F(\tau_i)$ = Integration($\mathfrak{g}^{\frac{1}{2}}(\tau_i)$, $[0,\tau_i]$)\; \label{algo:PRIMP:gora:F_tau}
}
Normalize $F(\tau^*)$ = $\frac{F(\tau)}{F(1)}$\; \label{algo:PRIMP:gora:F_tau_normalized}
Invert $F(\tau^*)$ into $\tau^*(t)$ = Interpolation($F(\tau^*)$, $t$)\; \label{algo:PRIMP:gora:tau_opt}
Reparameterize as $g(\tau^*)$ = Interpolation($\tau^*$, $g(\tau)$)\; \label{algo:PRIMP:gora:x_opt}

\caption{Globally Optimal Reparameterization Algorithm (GORA) for $\SE(3)$ sequence}
\label{algo:PRIMP:gora}

\end{algorithm}

\subsection{Computation of relative pose distribution} \label{sec:PRIMP:relative_distribution}
For a set of poses $\left\{ g^{(k)}_i \right\}$ in $\SE(3)$ at each step $i$, the sample mean $\mu_i \in \SE(3)$ satisfies 
\begin{equation}
\sum_{k=1}^{m} \log(\mu^{-1}_i g^{(k)}_i) = \mathbb{O} \,,
\label{eq:sample_mean}
\end{equation} 
which can be iteratively solved \cite{ackerman2013probabilistic}. Here $\SE(3)$ is treated as a matrix Lie group and $\log(\cdot)$ is the matrix logarithm. The mean trajectory can be directly computed from the demonstration set. 

The initial covariance $\Sigma_{i,i+1}$ encodes the uncertainty of $(i+1)^{\rm th}$ step given the $i^{\rm th}$ step. It is estimated by the set of relative poses, \ie $\left\{ \Delta^{(k)}_{i,i+1} = \left( g^{(k)}_i \right)^{-1} g^{(k)}_{i+1} \right\}$. With this set, the sample covariance can be computed as 
\begin{equation}
\Sigma_{i,i+1} = \frac{1}{m} \, \sum_{k=1}^{m} \log^{\vee}\left( \mu^{-1}_{i,i+1} \Delta^{(k)}_{i,i+1} \right) \, \log^{\vee \, T}\left( \mu^{-1}_{i,i+1} \Delta^{(k)}_{i,i+1} \right) \,,
\label{eq:sample_rel_covariance}
\end{equation}
where $\mu_{i,i+1}$ can be computed using Eq. \eqref{eq:sample_mean} but with the relative poses $\Delta^{(k)}_{i,i+1}$ as inputs, and the $\vee$ operator extracts the Lie algebra coefficients into a vector (as defined in \cite{chirikjian2011stochastic}).

\subsection{Probabilistic encoding of joint distribution on SE(3) trajectories} \label{sec:PRIMP:joint_distribution}
After computing the trajectory distribution with mean $\{ \mu_0, \mu_1,...,\mu_n \}, \, \mu_i \in \SE(3)$ and covariance between adjacent steps $\{ \Sigma_{0,1}, \Sigma_{1,2},...,\Sigma_{n-1,n} \}, \, \Sigma_{i,i+1} \in \IR^{6 \times 6}$ from Sec. \ref{sec:PRIMP:relative_distribution}, the joint distributions of the whole trajectory can be computed. The idea is to learn the distribution locally with end constraints. The solution is inspired by the concept of \textit{loop entropy} \cite{chirikjian2011modeling}, in which the probabilitic description of ensembles of end-constrained $\SE(3)$ paths was formulated. Assuming the variation of $i^{\rm th}$ pose only depends on its two neighboring poses and $g_0 = \mu_0$ is fixed, the joint probability density can be expressed as
\begin{equation}
\rho(g_1, g_2,...,g_{n}) = \prod_{i=0}^{n-1} \rho(g_{i+1} | g_i) \,,
\label{eq:condition_pdf}
\end{equation}
where $\rho(g_{i+1} | g_i)$ is the conditional probability of the $(i+1)^{\rm th}$ pose given the $i^{\rm th}$ pose.

If the intermediate steps along the trajectory are subject to Gaussian distributions with small variations, explicit results can be shown as 
\begin{equation}
\rho(g_1, g_2,...,g_{n}) = \eta \exp \left(-\frac{1}{2} \xx^T_{1,...,n} \Sigma^{'-1}_{1,...,n} \xx_{1,...,n} \right) \,,
\label{eq:conditional_pdf_gaussian_full}
\end{equation}
where $\exp(\cdot)$ here is the scalar exponential, $\eta = (2 \pi)^{-3 (n-1)} |\det \Sigma'_{1,...,n}|^{-\frac{1}{2}}$ is the normalizing constant and
$$
\xx_{1,...,n} \doteq [\xx_1^T,..., \xx_i^T,..., \xx_{n}^T]^T \,,
$$ 
where $\xx_i = \log^{\vee}(\mu_i^{-1} g_i)$. The non-zero elements of $\Sigma'^{-1}_{1,...,n}$ are
\begin{equation}
\begin{aligned}
\Sigma'^{-1}_{1,...,n}(i,i) &= \left\{ \begin{aligned}
& \Sigma^{-1}_{i-1,i} + \widetilde{\Sigma}^{-1}_{i,i+1} & (i = \{1,...,n-1\}) \\
& \Sigma^{-1}_{i-1,i} & (i = n)
\end{aligned} \right. \\
\Sigma'^{-1}_{1,...,n}(i,i+1) &= -Ad^{-T}_{i,i+1} \Sigma^{-1}_{i,i+1} \,\, (i = \{1,...,n-1\}) \\
\Sigma'^{-1}_{1,...,n}(i+1,i) &= -\Sigma^{-1}_{i,i+1} Ad^{-1}_{i,i+1} \,\, (i = \{1,...,n-1\}) \,,
\end{aligned}
\label{eq:condition_pdf_joint_cov}
\end{equation}
where $\Sigma'_{1,...,n} \in \IR^{6n \times 6n}$. $Ad(g)$ is the adjoint operator for $g$ in a Lie group, which is defined as 
$$
Ad(g) \widehat{\xx} \doteq g \widehat{\xx} g^{-1} \,,
$$
where $\widehat{\cdot}$ is the inverse operation of $\vee$ for elements in Lie algebra and 
$$
Ad_{i, i+1} \doteq Ad(\mu^{-1}_i \mu_{i+1})
$$ 
is the adjoint operator for the relative poses between $\mu_i$ and $\mu_{i+1}$ (the explicit expression of adjoint operator for $\SE(3)$ can be referred to Appendix \ref{appendix:se}) and
\begin{equation}
\widetilde{\Sigma}_{i,i+1} = Ad_{i,i+1} \Sigma_{i,i+1} Ad^{T}_{i,i+1} \,.
\label{eq:condition_pdf_joint_cov:tilde_cov}
\end{equation}
In this formulation, the variable $\xx_{1,...,n} \sim \mathcal{N}({\bf 0}, \Sigma'_{1,...,n})$ is interpreted as a Gaussian with zero mean. Detailed derivations can be referred to Appendix \ref{appendix:joint_pdf}.

\subsection{Adaptation to novel situations} \label{sec:PRIMP:adaptation}
One of the most essential abilities of an LfD method is its adaptability to novel unseen situations. The adaptation to via poses in $\SE(3)$ is firstly discussed. Then, its equivariant property is studied when there is a change of viewing frame. Finally, the learned distribution is fused with the robot-specific workspace density.

\subsubsection{Adaptation to via poses with uncertainties} \label{sec:PRIMP:adaption:via_point}
\begin{figure}[t]
\centering
\subfloat[Original trajectory mean and samples]{\includegraphics[scale=0.32, trim = 100 20 80 20, clip]{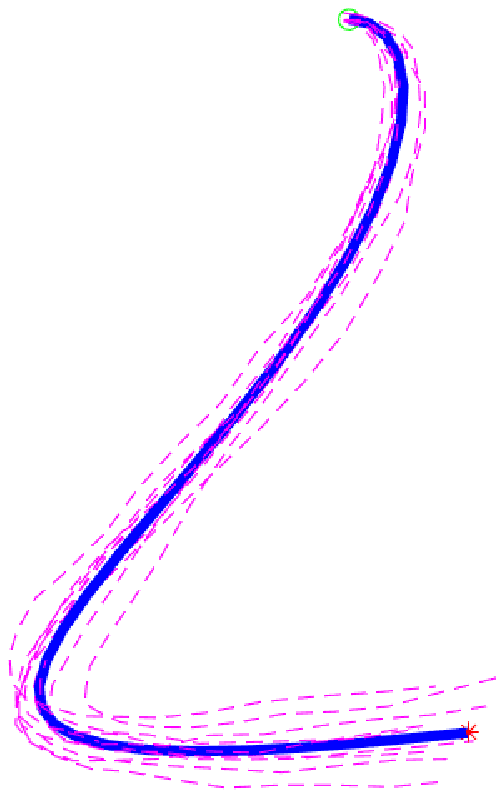}}
~
\subfloat[Adaptation to a new goal pose]{\includegraphics[scale=0.32, trim = 100 20 80 20, clip]{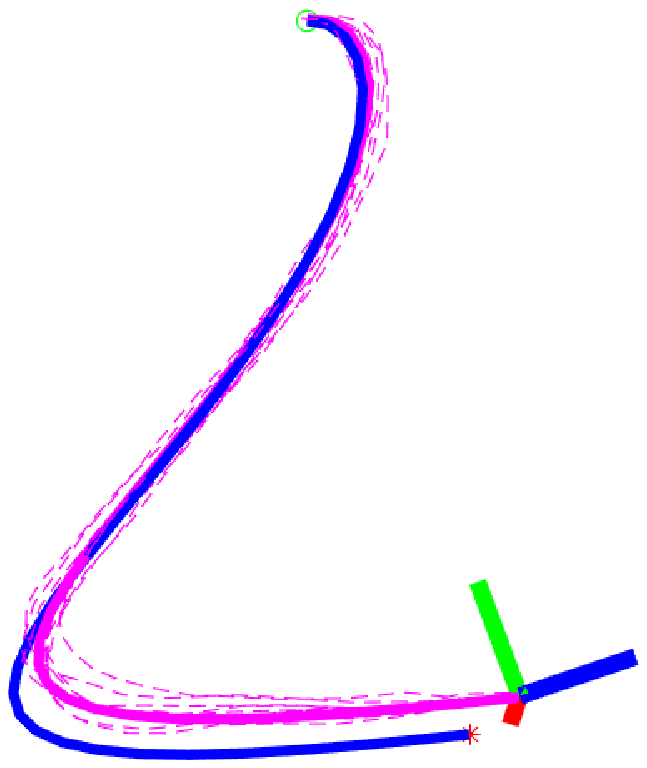}}
~
\subfloat[Adaptation to new via poses]{\includegraphics[scale=0.32, trim = 100 20 80 20, clip]{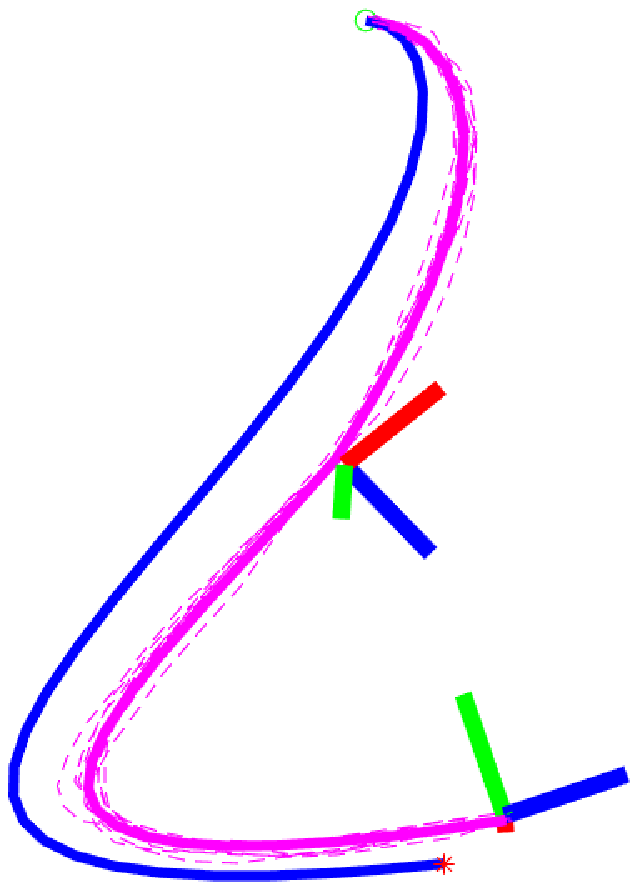}}
\caption{Examples of the adaptation to new via poses with uncertainties. The solid blue and magenta curves are the means of the encoded joint prior (Eq. \eqref{eq:condition_pdf_joint_cov}) and posterior (Eq. \eqref{eq:condition_pdf_via_point}) distribution, respectively. Dashed magenta curves are the random trajectory samples from the probability distribution.}
\label{fig:result:ex_condition_pdf_via_pose_observation}
\end{figure}

Suppose that the robot is asked to pass a via pose $g^*_i \in \SE(3)$ with uncertainty, which is described by covariance matrix $\Sigma^*_i$. The posterior distribution can be computed as follows. Using an observation model, the new via pose be defined as a function of the variable $\xx_i$, \ie
\begin{equation}
g^*_i \doteq \mu_i \exp \left(\widehat{\xx}_i\right) \exp \left(\widehat{\xxi}\right) = \mu_i \exp \left(\widehat{C_i \xx}_{1,...,n}\right) \exp \left(\widehat{\xxi}\right) \,,
\label{eq:observation_mean}
\end{equation}
where $\exp(\widehat{\cdot})$ here is the matrix exponential that maps from Lie algebra to Lie group, $\exp \left(\widehat{\xxi}\right) \sim \mathcal{N}(\II, \Sigma^*_i)$ is subject to the desired covariance, 
$$
C_i = {\bf e}^T_i \otimes \II_6
$$ 
indicates the block of variables for the $i^{\rm th}$ step and $\otimes$ denotes the Kronecker product. Defining a new variable, \ie $\yy \doteq \log^{\vee}\left( \mu_i^{-1} g^*_i \right)$, gives
\begin{equation}
\yy = \log^{\vee} \left( \exp \left(\widehat{C_i \xx}_{1,...,n}\right) \exp \left(\widehat{\xxi}\right) \right) \approx C_i \xx_{1,...,n} + \xxi \,.
\label{eq:observation_model}
\end{equation}
Using Eq. \eqref{eq:observation_model}, the mean and covariance of the posterior distribution can be computed as
\begin{equation}
\begin{aligned}
K_i & \doteq \Sigma'_{1,...,n} C^T_i \left( C_i \Sigma'_{1,...,n} C^T_i + \Sigma^*_i \right)^{-1} \\
\hat{\xx}_{1,...,n} &= K_i \log^{\vee}\left( \mu^{-1}_i g^*_i \right) \\
\hat{\Sigma}'_{1,...,n} &= (\II - K_i C_i) \Sigma'_{1,...,n} \,.
\end{aligned}
\label{eq:condition_pdf_via_point}
\end{equation}
Figure \ref{fig:result:ex_condition_pdf_via_pose_observation} shows the adaptation to novel uncertain via poses.

\subsubsection{Equivariant adaptation to the change of view} \label{sec:PRIMP:adaptation:change_view}
To change the viewing frame, a group action is applied. Suppose $h \in \SE(3)$ is the relative transformation from the current frame ($O$) to a new frame ($A$), then the pose $g$ viewed in frame $O$ can be switched to be viewed in frame $A$ as $g^{o} = h^{-1} g h$ \cite{chirikjian2018pose}.

The conditional probability between two adjacent frames after the change of view can be computed as 
\begin{equation}
\xx^{o}_i = \log^{\vee}(h^{-1} \mu^{-1}_i g_i h) = Ad^{-1}(h) \xx_i \,.
\end{equation}
The same expression can be obtained for $\xx^{o}_{i+1}$. Then, the joint variable $\xx_{i,i+1} = [\xx^T_i, \xx^T_{i+1}]^T$ has the expression after the change of view as 
\begin{equation}
\xx^{o}_{i,i+1} = \left(\begin{matrix}
Ad^{-1}(h) & \mathbb{O}_{6 \times 6} \\
\mathbb{O}_{6 \times 6} & Ad^{-1}(h)
\end{matrix}\right) \xx_{i,i+1} \,.
\label{eq:change_view_joint_variable}
\end{equation}

For a Gaussian distribution, explicitly writing down the quadratic term in the exponent gives
\begin{equation}
\rho(g_{i+1} | g_i) \propto \exp \left( -\frac{1}{2} \left\| \xx^{o}_{i+1} - Ad^{o}_{i,i+1} \xx^{o}_i \right\|^2_{\Sigma^{o \, -1}_{i,i+1}} \right) \,,
\label{eq:change_view_joint_gaussian}
\end{equation}
where $\Sigma^{o}_{i,i+1} = Ad^{-1}(h) \Sigma_{i,i+1} Ad^{-T}(h)$, $Ad^{o}_{i,i+1} \doteq Ad(\mu^{o \, -1}_i \mu^{o}_{i+1})$, $\| \xx \|^2_W = \xx^T W \xx$ for a vector $\xx \in \IR^n$ and $-T$ denotes the inverse of the transpose of a matrix. Detailed derivation can be found in Appendix \ref{appendix:equivariance}. From Eq. \eqref{eq:change_view_joint_gaussian}, the conditional probability viewed in frame $A$ is also a Gaussian with zero mean and covariance $\Sigma^{o}_{i,i+1}$. Therefore, for the joint variable $\xx_{i,i+1}$, we can compute the inverse of joint covariance in the new frame as
\begin{equation}
\Sigma'^{o \, -1}_{i,i+1} = \left(\begin{matrix}
Ad^{o \, -T}_{i,i+1} \Sigma^{o \, -1}_{i,i+1} Ad^{o \, -1}_{i,i+1} & -Ad^{o \, -T}_{i,i+1} \Sigma^{o \, -1}_{i,i+1} \\
-\Sigma^{o \, -1}_{i,i+1} Ad^{o \, -1}_{i,i+1} & \Sigma^{o \, -1}_{i,i+1}
\end{matrix}\right) \,.
\label{eq:change_view_joint_gaussian_cov}
\end{equation}


From the above derivations, 
\begin{equation}
f(h \circ \xx_{i,i+1}) = h \odot f(\xx_{i,i+1}) \,,
\label{eq:primp:adaptation:equivariance}
\end{equation}
where $h \circ \xx_{i,i+1}$ has the form in Eq. \eqref{eq:change_view_joint_variable} and $h \odot f(\xx_{i,i+1}) \sim \mathcal{N}({\bf 0}, \Sigma'^{o}_{i,i+1})$ with the covariance from Eq. \eqref{eq:change_view_joint_gaussian_cov}, which shows the equivariance of the conditional distribution under the change of view.

\begin{figure}[t]
\centering
\subfloat[Encoded joint distribution.]{\includegraphics[trim = 50 0 50 0, clip, scale=0.35]{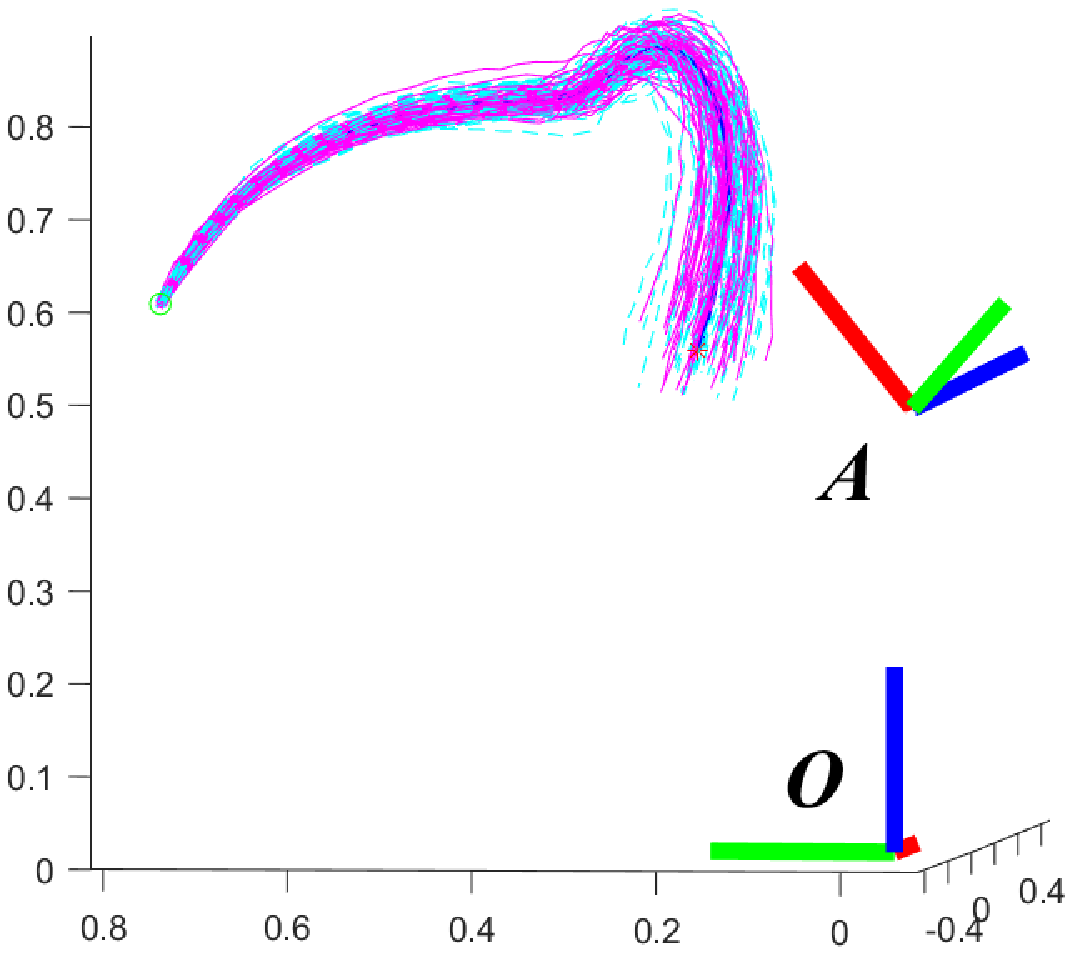} \label{fig:result:ex_equivariance:encoding}}
~
\subfloat[Distribution after conditioning on a via pose.]{\includegraphics[trim = 50 0 50 0, clip, scale=0.35]{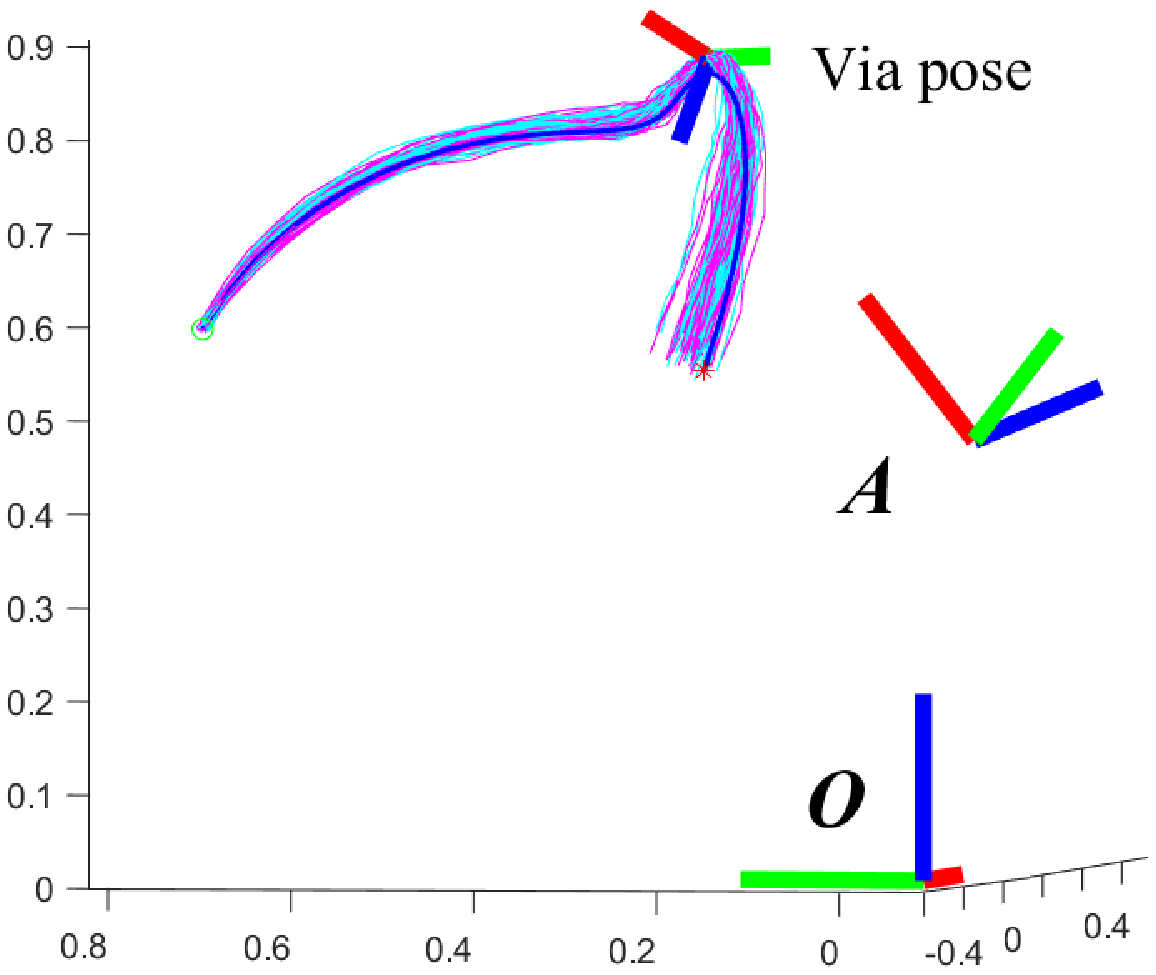} \label{fig:result:ex_equivariance:via_samples}}
\caption{Equivariance property under the change of viewing frame. The samples as viewed in the original frame $O$ are shown in magenta, and samples after the change of view using Eq. \eqref{eq:change_view_joint_gaussian_cov_traj} are shown in cyan (trajectories are sampled as viewed in the new frame $A$ but transformed back to frame $O$ for each pose for illustration purpose).}
\label{fig:result:ex_equivariance}
\end{figure}

Using the same derivation process, the distribution of the whole trajectory can be obtained. The variable becomes
\begin{equation}
\xx^{o}_{1,...,n} = \left( \II_n \otimes Ad^{-1}(h) \right) \xx_{1,...,n} \,,
\label{eq:change_view_joint_variable_traj}
\end{equation}
where $\xx_{1,...,n}$ is defined in Eq. \eqref{eq:conditional_pdf_gaussian_full}. The distribution has zero mean and covariance, which has the same structure as Eq. \eqref{eq:condition_pdf_joint_cov}. The difference is for Eq. \eqref{eq:condition_pdf_joint_cov:tilde_cov}, which becomes
\begin{equation}
\begin{aligned}
\Sigma^{o}_{i,i+1} &= Ad^{-1}(h) \Sigma_{i,i+1} Ad^{-T}(h) \,, \text{and} \\
\widetilde{\Sigma}^{o}_{i,i+1} &= Ad^{o}_{i,i+1} \Sigma^{o}_{i,i+1} Ad^{o \, T}_{i,i+1} \,.
\end{aligned}
\label{eq:change_view_joint_gaussian_cov_traj}
\end{equation}
Equations \eqref{eq:change_view_joint_variable_traj} and \eqref{eq:change_view_joint_gaussian_cov_traj} also satisfy the equivariance property (as in Eq. \eqref{eq:primp:adaptation:equivariance}). Demonstrations on the equivariance property for the change of viewing frame are shown in Fig. \ref{fig:result:ex_equivariance}. 


\subsubsection{Adaptation to robot-specific workspace density} \label{sec:PRIMP:adaptation:wd}
PRIMP learns in the workspace of the robot instead of joint space, which is robot-agnostic and makes skill transfer among different robots easy. However, an important issue to consider is the adaptation to robot-specific workspace limits and reachability. Previous work has extensively investigated the density of the robot workspaces, in which the more reachable space of the end effector has a higher probability \cite{chirikjian2000engineering,wang2008nonparametric}. It has been widely used to efficiently solve the inverse kinematics of a serial hyper-redundant manipulator \cite{ebert1996inverse,chirikjian1998numerical,wang2004workspace}. In this work, the same concept is applied but in a different way. The workspace density is used to inform the learned trajectory distribution, in order to stay closer to the higher probability region within the robot workspace. The following introduces how to fuse the learned distribution with the workspace density of a specific robot, so as to maximize the mobility when moving along the learned trajectory.

The mathematical foundation is based on the convolution on $\SE(3)$, \ie 
$$
(\rho_1 * \rho_2)(g) \doteq \int_G \rho_1(h) \rho_2(h^{-1} \circ g) dh \,,
$$
where $h$ is a dummy variable and $dh$ is the Haar measure on $\SE(3)$ \cite{chirikjian2000engineering}. For a serial manipulator with $m$ joints, the workspace density is computed as the convolution of Gaussian distributions, \ie 
$$
\rho_{\rm wd}(g) = (\rho_1 * ... * \rho_m)(g) \,,
$$
where $\rho_i(g) \,, i=1,...,m$ is the Gaussian distribution of all possible poses of the distal end of the $i^{\rm th}$ link. In practice, a discrete set of joint angles is sampled uniformly. And the pose of the distal end is computed and approximated as a Gaussian defined by sample mean and covariance using Eqs. \eqref{eq:sample_mean} and \eqref{eq:sample_rel_covariance}. The resulting workspace density is also a Gaussian in $\SE(3)$, whose mean can be computed as 
$$
g_{\rm wd} = \mu_1 \, \mu_2 \, ... \, \mu_m \,.
$$
The covariance $\Sigma_{\rm wd}$ can be approximated by an iterative process of two-fold convolution, \ie 
$$
\Sigma_{1*2} = Ad(\mu^{-1}_2) \, \Sigma_1 \, Ad^{T}(\mu^{-1}_2) + \Sigma_2 \,,
$$
which is the first-order approximation of the covariance \cite{smith2003computing,wang2008nonparametric,barfoot2014associating}.

Then, the distribution of each intermediate pose along the trajectory is conditioned by this density function. The idea is analogous to that in Sec. \ref{sec:PRIMP:adaption:via_point}, in which each intermediate pose is asked to pass through the desired $g_{\rm wd}$ with uncertainty $\Sigma_{\rm wd}$. For each step $i$, the workspace density can be approximated as
\begin{equation}
g_{\rm wd} \doteq \mu_i \exp \left(\widehat{\xx}_i \right) \exp \left( \widehat{\xxi} \right) \,,
\end{equation}
where $\exp \left(\widehat{\xxi}\right) \sim \mathcal{N}(\II, \Sigma_{\rm wd})$. Let $\yy_i \doteq \log^{\vee}\left( \mu_i^{-1} g_{\rm wd} \right)$, then 
$$
\yy_i \approx \xx_i + \xxi \,.
$$
Stacking the variables for all steps together gives 
$$
\yy = [\yy^T_1,..., \yy^T_n]^T \,.
$$
Therefore, the trajectory distribution after fusing with the workspace density can be computed by
\begin{equation}
\begin{aligned}
K & \doteq \Sigma'_{1,...,n} \left( \Sigma'_{1,...,n} + \II_{n} \otimes \Sigma_{\rm wd} \right)^{-1} \\
\hat{\xx}_{1,...,n} &= K \yy \\
\hat{\Sigma}'_{1,...,n} &= (\II - K) \Sigma'_{1,...,n} \,.
\end{aligned}
\label{eq:fusion_workspace_density}
\end{equation}

\begin{figure}
\centering
\subfloat[Franka Emika Panda]{\includegraphics[scale=0.4, trim=60 0 60 30, clip]{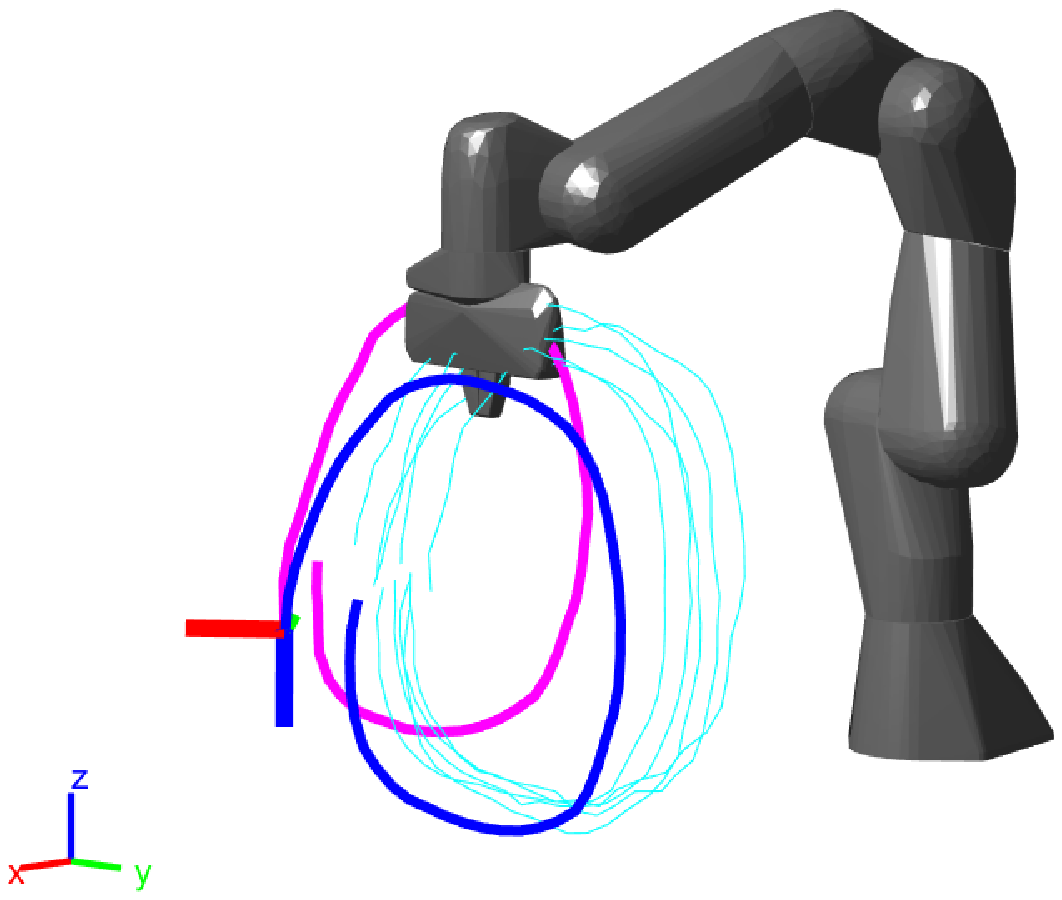}}
~
\subfloat[Kinova Gen3]{\includegraphics[scale=0.4, trim=60 0 60 30, clip]{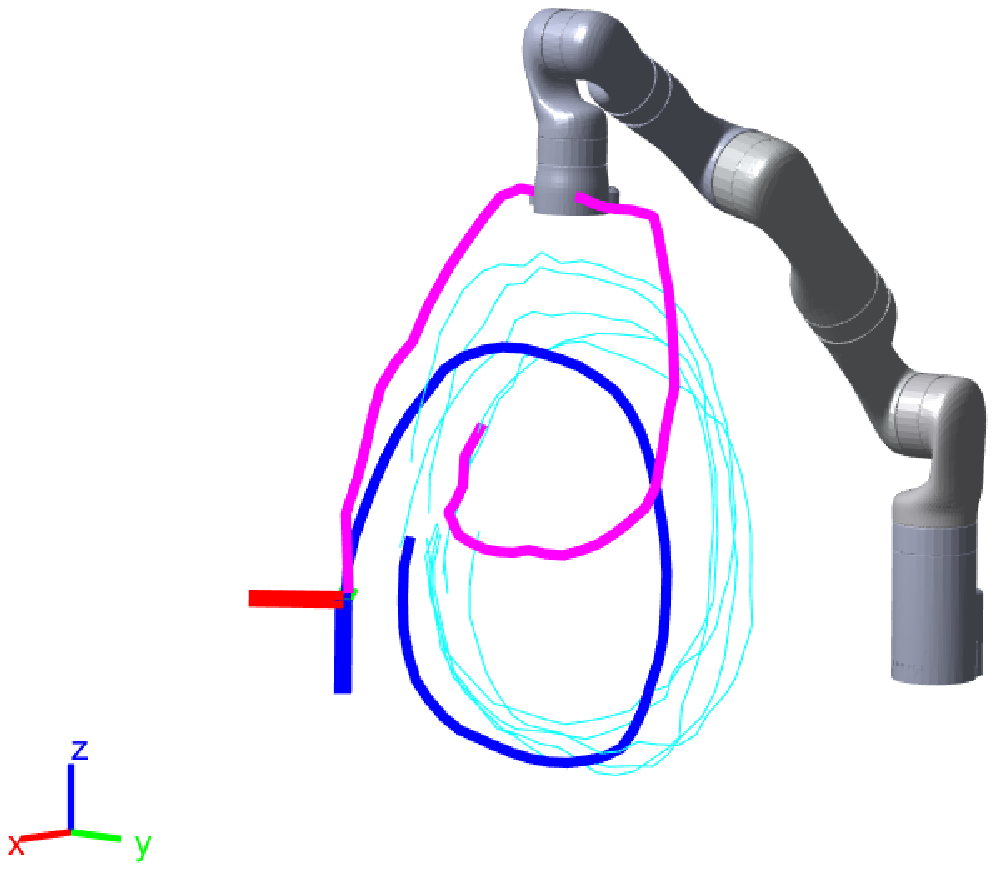}}
\caption{Fusion with robot-specific workspace density. Thin cyan curves are the demonstrated trajectories using Franka robot; solid thick blue and magenta curves are the mean trajectory without and with the fusion, respectively. The end effector is placed at a random intermediate step along the fused trajectory mean.}
\label{fig:PRIMP:with_wd}
\end{figure}

Figure \ref{fig:PRIMP:with_wd} demonstrates the fusion with robot-specific workspace density. The demonstrations are conducted using Franka Emika Panda robot. Then, PRIMP is fused by the workspace density of both Panda and Kinova Gen3 robots, followed by adapting to a via pose.

\subsection{Extensions to the Pose Change Group (PCG)} \label{sec:PRIMP:pcg}
\begin{figure}[t]
\centering
\subfloat[$\SE(3)$]{\includegraphics[scale=0.35, trim=40 0 30 30, clip]{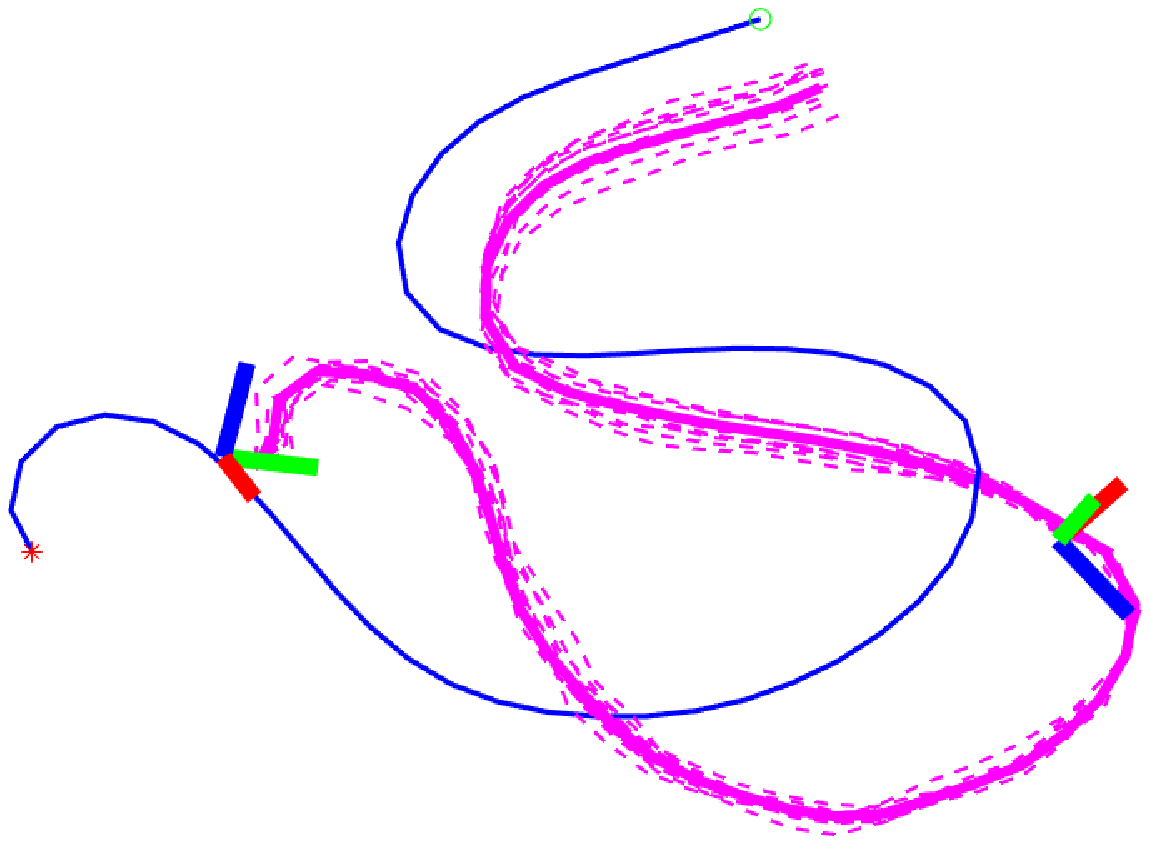}}
~
\subfloat[$\PCG(3)$]{\includegraphics[scale=0.35, trim=40 0 30 30, clip]{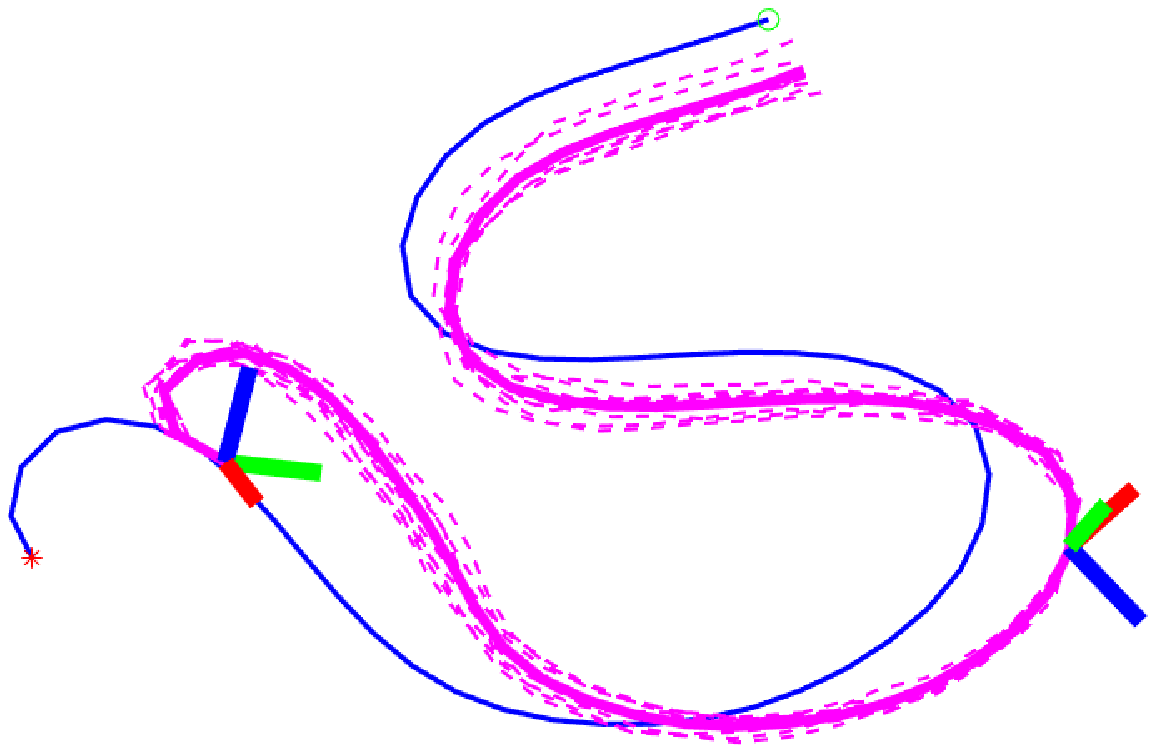}}
\caption{PRIMP modeled on $\SE(3)$ and $\PCG(3)$. Solid blue and magenta curves are the mean of the demonstrated and conditioned trajectories, respectively.}
\label{fig:PRIMP:ex_condition_pdf_goal_pose_se_pcg}
\end{figure}

In addition to $\SE(3)$, the trajectory can also be represented by $\PCG(3)$ \cite{chirikjian2018pose}, where the local variable for each frame becomes
$
\xx_i \doteq \left(\begin{matrix}
\log^{\vee}(Q_i^T R_i) \\
{\bf t}_i - {\bf s}_i \end{matrix}\right)
$, where $Q_i$ and $s_i$ are the means of rotation and translation part, respectively. Using the new local variable and $\PCG(3)$ operations, Eqs. \eqref{eq:condition_pdf_via_point}, \eqref{eq:sample_mean} and \eqref{eq:sample_rel_covariance} are changed accordingly. A qualitative comparison for the $\SE(3)$ and $\PCG(3)$ trajectories is demonstrated in Fig. \ref{fig:PRIMP:ex_condition_pdf_goal_pose_se_pcg}.

\section{Motion planning guided by PRIMP} \label{sec:wstomp}
This section introduces \textit{Workspace-STOMP}, a novel guided motion planning algorithm using the trajectory distribution computed by PRIMP. A cost function for the end effector trajectory is proposed to guide the STOMP algorithm. The cost is computed based on the distance metric in $\SE(3)$ between each rollout trajectory at each iteration and the workspace trajectory distribution learned by PRIMP.

At each iteration, STOMP defines a set of random samples in joint space, each of which is denoted as a ``rollout'', \ie ${\bf q}$. To compute the distance metric of each rollout with the learned trajectory distribution, the trajectory of the end effector is computed via forward kinematics, denoted as $g({\bf q}, t) \in \SE(3) \times \mathcal{T}$. Then, a number of $m_r$ random samples from the reference trajectory distribution are generated, denoted as $g^{(k)}_r = \left(R^{(k)}_r, {\bf t}^{(k)}_r\right)$. And the cost function ${\bf c}({\bf q}_i, t_i)$ for $i^{\rm th}$ time step is computed as
\begin{equation}
\begin{aligned}
{\bf c}({\bf q}_i, t_i) =& \frac{1}{m_{\rm r}} \, \sum_{k=1}^{m_{\rm r}} \left( w_{\rm rot} \left\| \log^{\vee} \left( R^{T}({\bf q}_i, t_i) \, R^{(k)}_{\rm r}(t_i) \right) \right\| \right. \\
& \left. + w_{\rm tran} \left\|{\bf t}({\bf q}_i, t_i) - {\bf t}^{(k)}_{\rm r}(t_i) \right\| \right)
\end{aligned} \,.
\label{eq:wstomp:cost}
\end{equation}
The weights $w_{\rm rot}$ and $w_{\rm tran}$ for rotation and translation parts in the distance function are set by users, which by default are both 1.0. The computational process is shown in Alg. \ref{algo:wstomp:cost_function}. 

\begin{algorithm}[t]
\SetAlgoLined
\SetKwInOut{Param}{Parameters}
\SetKwInOut{In}{Inputs}
\SetKwInOut{Out}{Outputs}

\In{${\bf q}$: Rollout joint angles;\\
$\left\{ g^{(k)}_{\rm r}(t) \right\}$: Sampled trajectories from the distribution learned by PRIMP}
\Param{$w_{\rm rot}$, $w_{\rm tran}$: Weights for rotation and translation parts in the distance function}
\Out{${\bf c}({\bf q}, t)$: Cost value for each rollout}

\For{time step $i$}{
    $g({\bf q}_i, t_i)$ = ForwardKinematics(${\bf q}_i$)\;
    Compute ${\bf c}({\bf q}_i, t_i)$ using Eq. \eqref{eq:wstomp:cost} \;
}

\caption{Cost function for Workspace-STOMP based on trajectory distribution}
\label{algo:wstomp:cost_function}

\end{algorithm}

\begin{figure}[t]
\centering
\subfloat[Letter $N$]{\includegraphics[trim = 180 0 150 0, clip, scale=0.12]{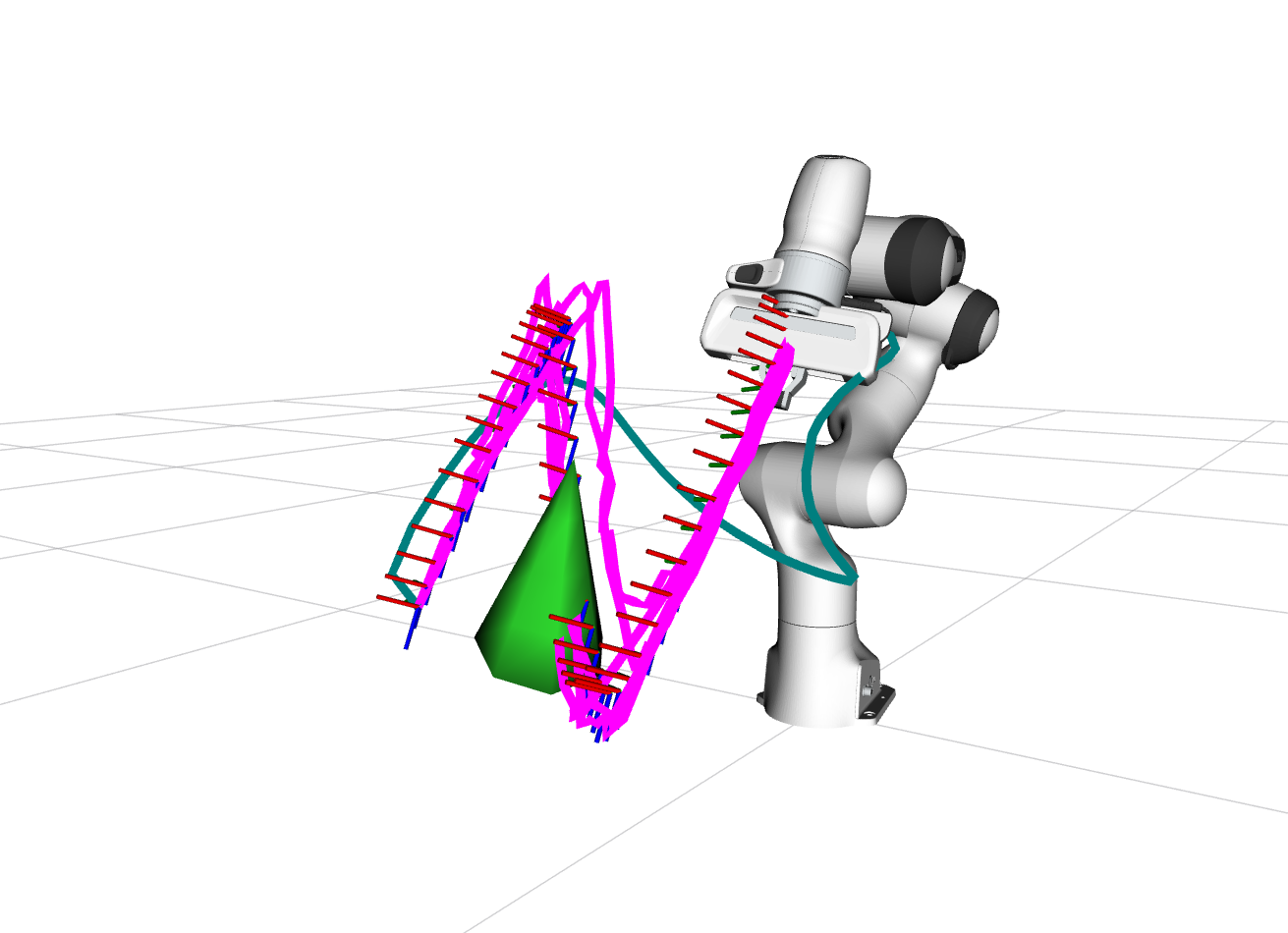}}
~
\subfloat[Letter $U$]{\includegraphics[trim = 150 0 150 0, clip, scale=0.12]{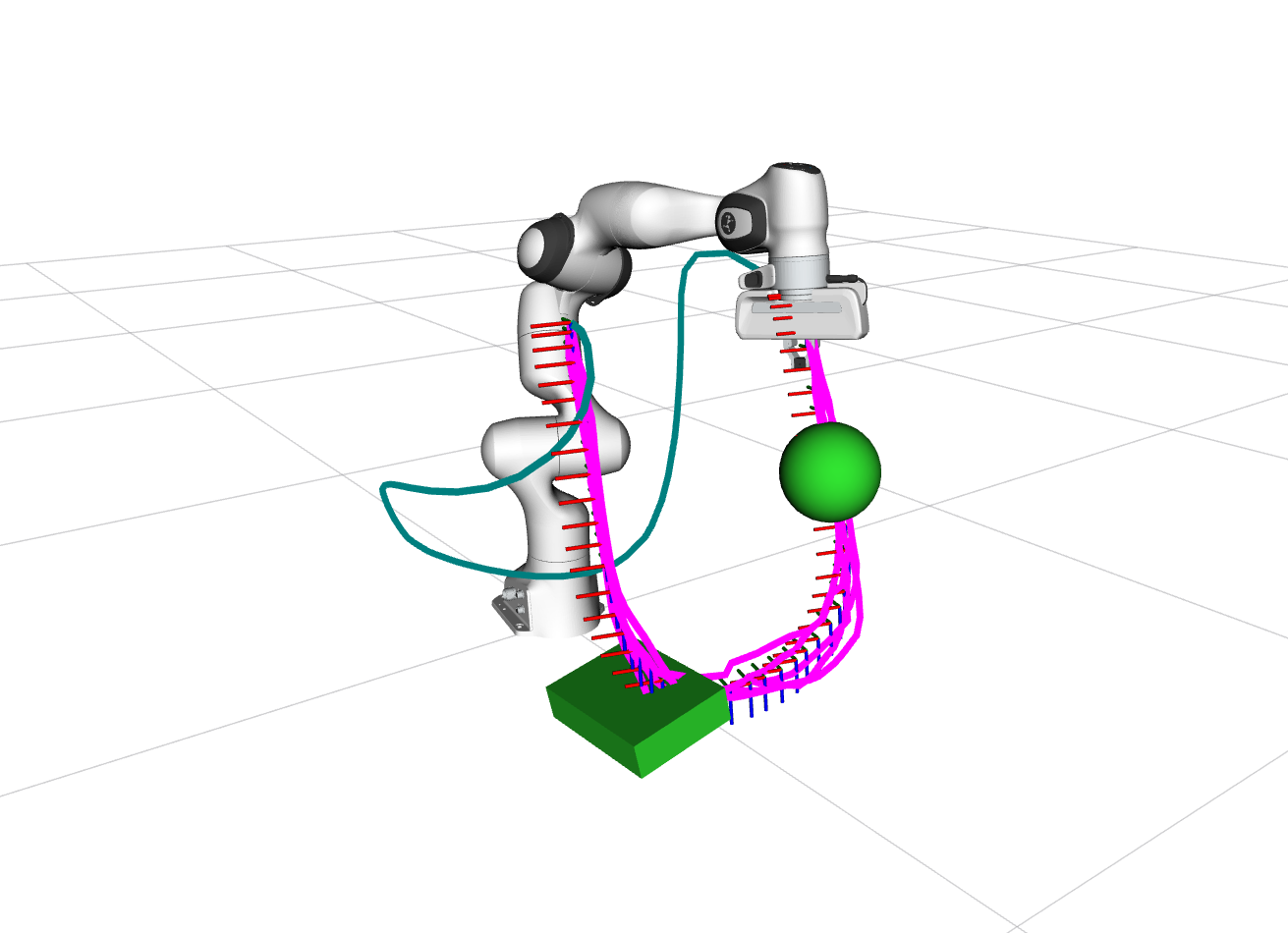}}
~
\subfloat[Letter $S$]{\includegraphics[trim = 150 0 150 0, clip, scale=0.12]{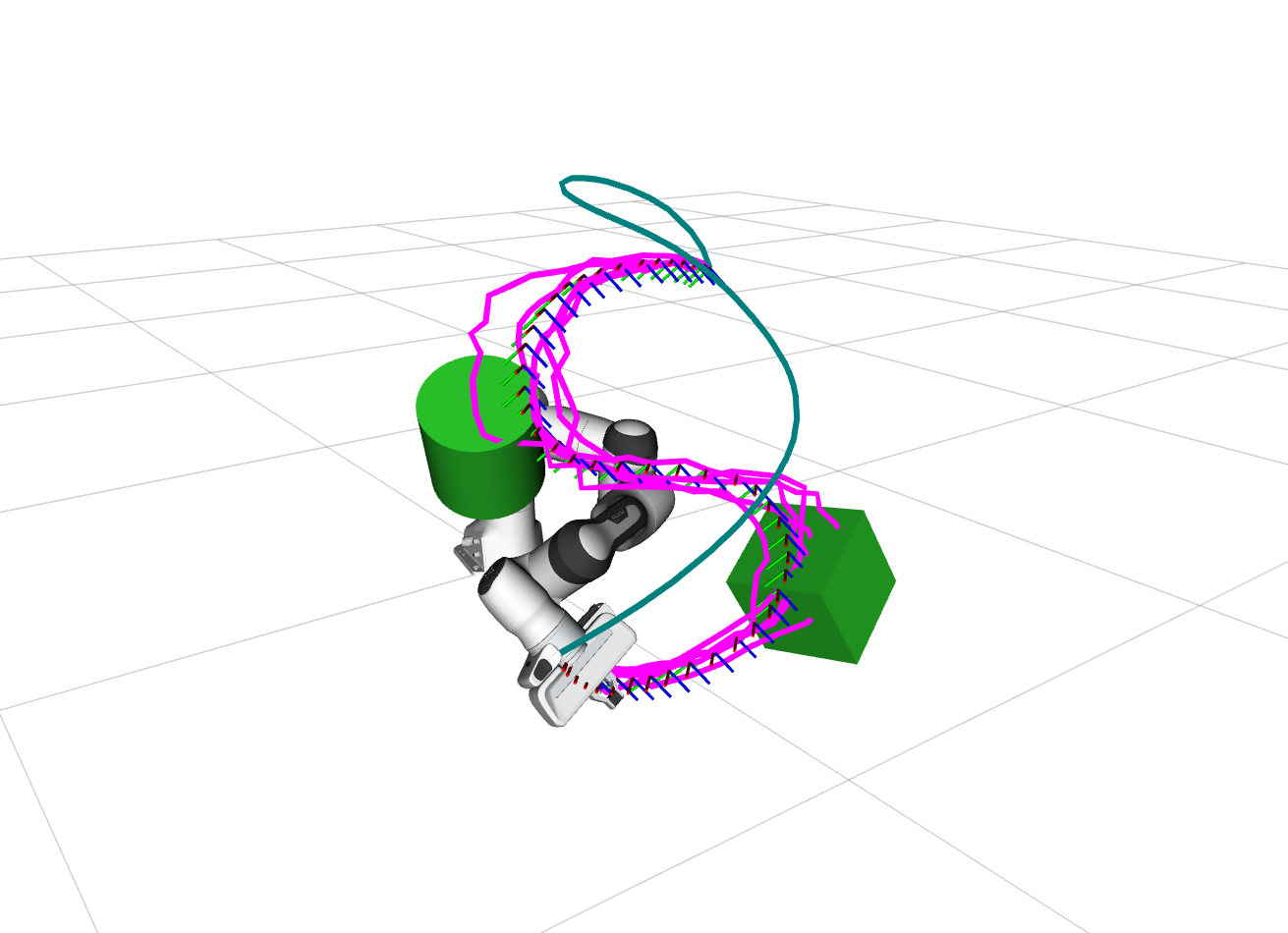}}
\caption{Planning using Workspace-STOMP to follow the learned trajectory distribution from PRIMP. Magenta lines are samples from posterior distribution; framed curves are mean trajectories; blue lines are the planned paths.}
\label{fig:wstomp:planning_results}
\end{figure}

The planner is initialized by the mean trajectory of the learned distribution. Then, a plug-in package of the proposed cost function is implemented in MoveIt! platform \cite{chitta2012moveit}. Simulations of writing letters ``N'', ``U'' and ``S'' using the proposed Workspace-STOMP are shown in Fig. \ref{fig:wstomp:planning_results}.

\section{Evaluations of the proposed method} \label{sec:benchmark}
The evaluations of the proposed PRIMP and Workspace-STOMP algorithms are conducted in this section: (1) PRIMP on $\SE(3)$ and $\PCG(3)$ as well as some popular state-of-the-art LfD methods (Sec. \ref{sec:benchmark:lfd}); (2) Properties of PRIMP in extrapolation and single-demonstration cases (Sec. \ref{sec:benchmark:PRIMP_property}); and (3) Variants of STOMP for motion planning (Sec. \ref{sec:benchmark:planner}).

\subsection{Comparison metrics}
\subsubsection{Similarity with original trajectory}
The averaged distance is computed between samples from the learned distribution and the demonstrated trajectories. Mathematically, the metric is defined as
\begin{equation}
D_{\rm demo} \doteq \frac{1}{m \, s \, N_{\rm step}} \sum_{k=1}^{m} \sum_{j=1}^{s} {\rm DTW} \left( \{g^{(j)}_{\rm learned}\}, \{g^{(k)}_{\rm demo}\} \right) \,,
\label{eq:benchmark:dist_demo}
\end{equation}
where $m, s$ are the numbers of originally demonstrated trajectories and sampled trajectories, respectively. ${\rm DTW}(\{\cdot\}, \{\cdot\})$ denotes the DTW distance between two sequences, where the distance metric in Lie group is used. In the benchmark experiments, the metrics for the rotation and translation parts are computed separately. In particular, $d_{\rm rot}(R_1, R_2) = \| R_1 - R_2 \|_F$ ($\| \cdot \|_F$ denotes the Frobenius norm) and $d_{\rm tran}({\bf t}_1, {\bf t}_2) = \| {\bf t}_1 - {\bf t}_2 \|_2$ ($\| \cdot \|_2$ denotes the 2-norm).

\subsubsection{Adaptability to new via pose}
The distance between the computed pose at a specific time step $i$ and the designated pose mean at the same step, \ie $\mu^*_i$, is computed as
\begin{equation}
D_{\rm via} \doteq \frac{1}{s} \sum_{j=1}^{s} d \left( g^{(j)}_{i}, \mu^*_i \right) \,.
\label{eq:benchmark:dist_via}
\end{equation}

\subsubsection{Similarity with reference trajectory}
The similarity of the computed trajectory with the reference trajectory is evaluated after motion planning. Given the reference end-effector trajectory $\{ g_{\rm ref, i} = (R_{\rm ref, i}, {\bf t}_{\rm ref, i}) \}$ and planned trajectory $\{ g_{\rm plan, i} = (R_{\rm plan, i}, {\bf t}_{\rm plan, i}) \}$, the error metric is defined as the accumulated distance between each step along the two trajectories, \ie
\begin{equation}
\begin{aligned}
e_{\rm rot} &= \sum_{i=1}^{N_{\rm step}} \| \log^{\vee}\left( R_{\rm ref, i}^T R_{\rm plan, i} \right) \|_2 \,, \\
e_{\rm tran} &= \sum_{i=1}^{N_{\rm step}} \| {\bf t}_{\rm ref, i} - {\bf t}_{\rm plan, i} \|_2 \,.
\end{aligned}
\label{eq:benchmark:e_ref}
\end{equation}

\subsection{Benchmarks among learning-from-demonstration methods} \label{sec:benchmark:lfd}
The data for LfD benchmarks include three types: (1) LASA hand-writing dataset \cite{khansari2011learning}; (2) end effector positions generated in simulation; and (3) real-world kinesthetic teaching for some common daily tasks. The former two types only include positional trajectories while the third type involves both position and orientation data. The existing probabilistic methods to be compared include: (1) ProMP \cite{paraschos2018using}, which only deals with translation part of the trajectories; and (2) Orientation-KMP \cite{huang2020toward}, which considers both translation and orientation parts. For Orientation-KMP method, the parameter for the kernel plays an essential role in encoding and generalizing the learned trajectory distribution. Therefore, the benchmarks consider different levels of magnitude to compare the performance.

\subsubsection{LASA handwriting dataset}
\begin{figure*}
\centering
\subfloat[$D_{\rm demo}$, translation part]{\includegraphics[scale=0.4, trim={20 140 20 140}, clip]{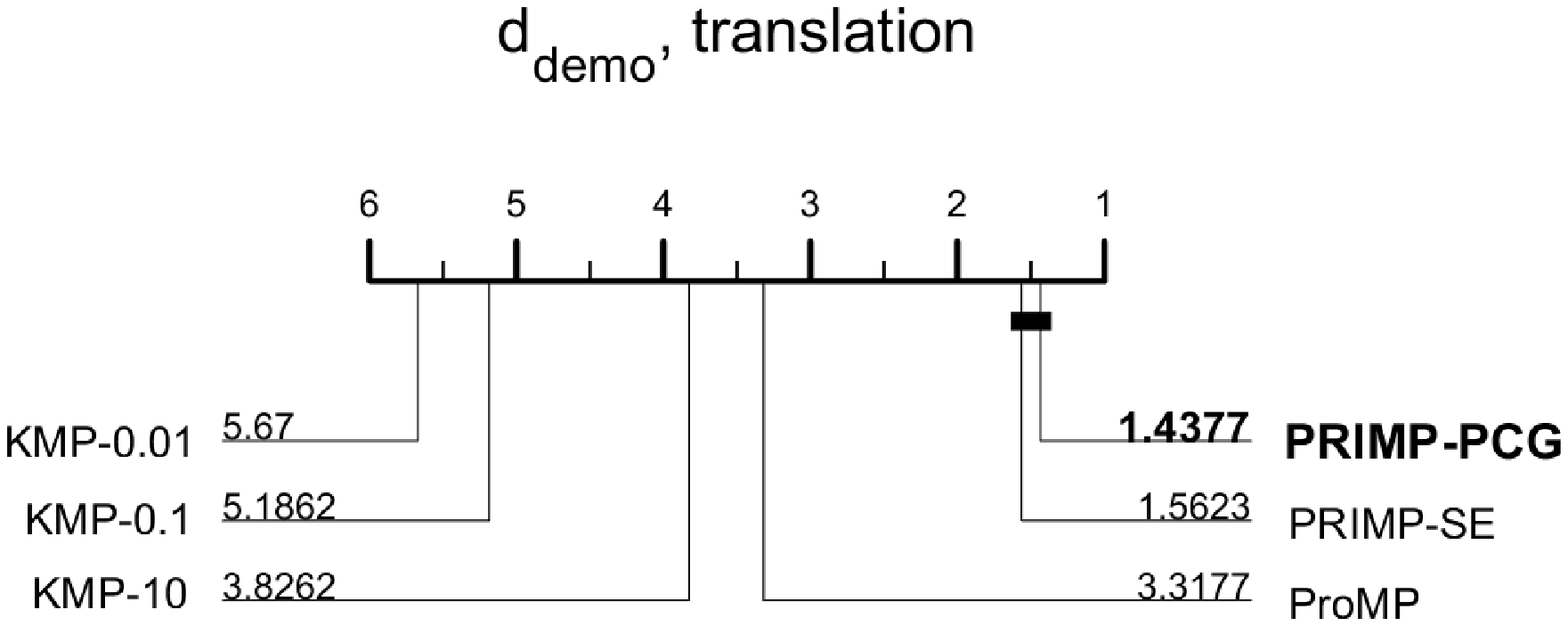}}
~
\subfloat[$D_{\rm via}$, translation part]{\includegraphics[scale=0.4, trim={20 140 20 140}, clip]{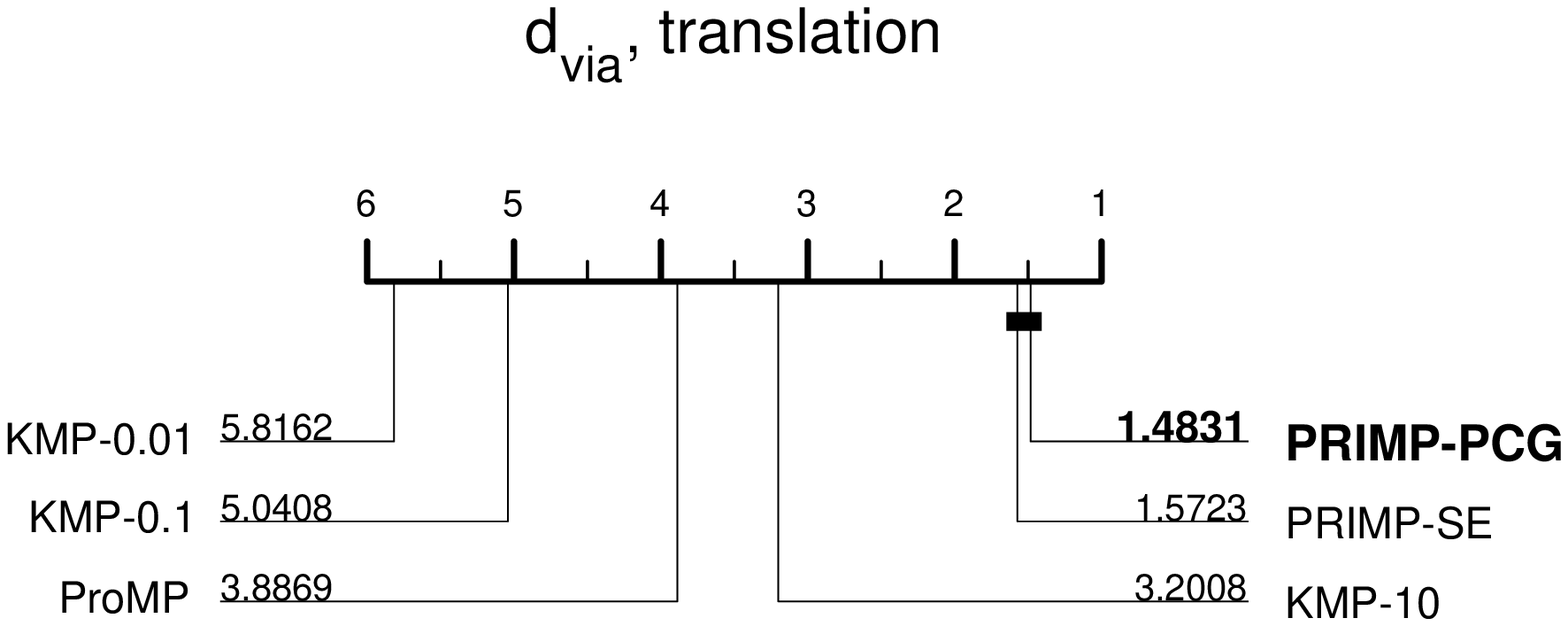}}
\caption{Critical difference diagram for performance ranking for the 26 categories in LASA handwriting dataset, with $\alpha=0.05$. The numbers are the averaged ranking of the methods, the smaller (further to the right) the better. The methods connected by bold lines do not have significant differences in performance.}
\label{fig:benchmark:lfd:lasa:cd}
\end{figure*}

This dataset involves 2D handwriting demonstrations. We then augment by adding a zero $z$-coordinate and identity 3D orientation for each pose in the trajectory. The 4 sets consisting of multi-modal data are omitted, so the comparisons are conducted in the remaining 26 data sets. The distance metrics in Eqs. \eqref{eq:benchmark:dist_demo} and \eqref{eq:benchmark:dist_via} are used to evaluate the performance. Since only translation is included in the demonstration, the evaluations only include the translation part. All the compared methods are ranked among each of the 26 categories. The averaged rankings are shown in a critical difference diagram in Fig. \ref{fig:benchmark:lfd:lasa:cd}.



\subsubsection{Demonstrations in simulation}
\begin{figure*}
\centering
\subfloat[$D_{\rm demo}$, ``Letter N'']{\includegraphics[scale=0.4, trim={0 0 0 25}, clip]{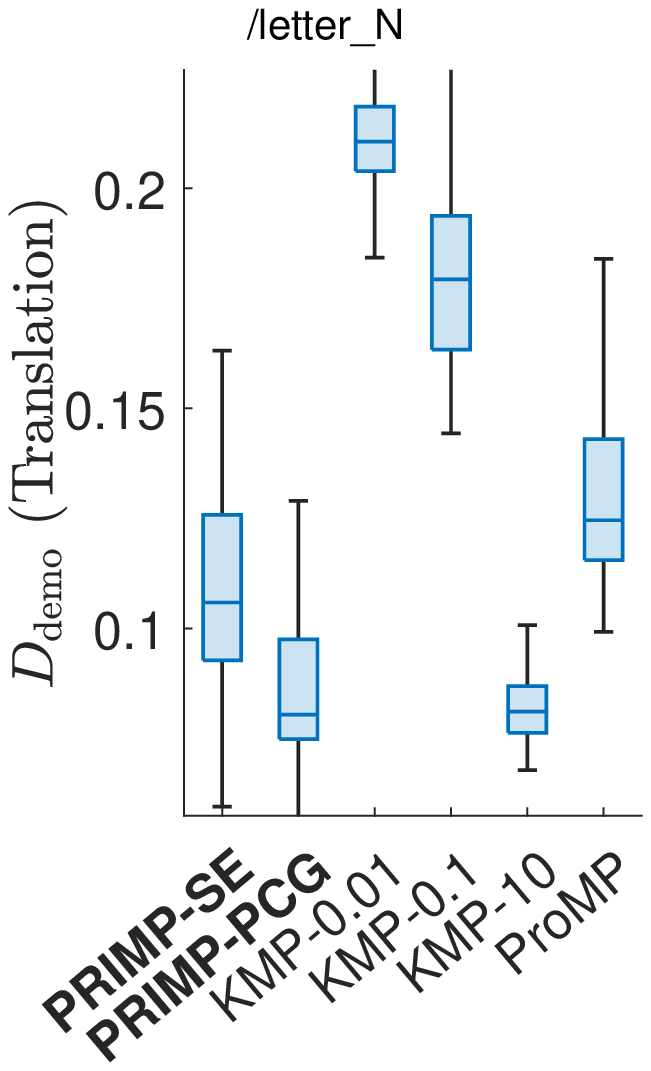} \label{fig:benchmark:lfd:simulation:distance:demo:letter_N}}
~
\subfloat[$D_{\rm demo}$, ``letter U'']{\includegraphics[scale=0.4, trim={0 0 0 25}, clip]{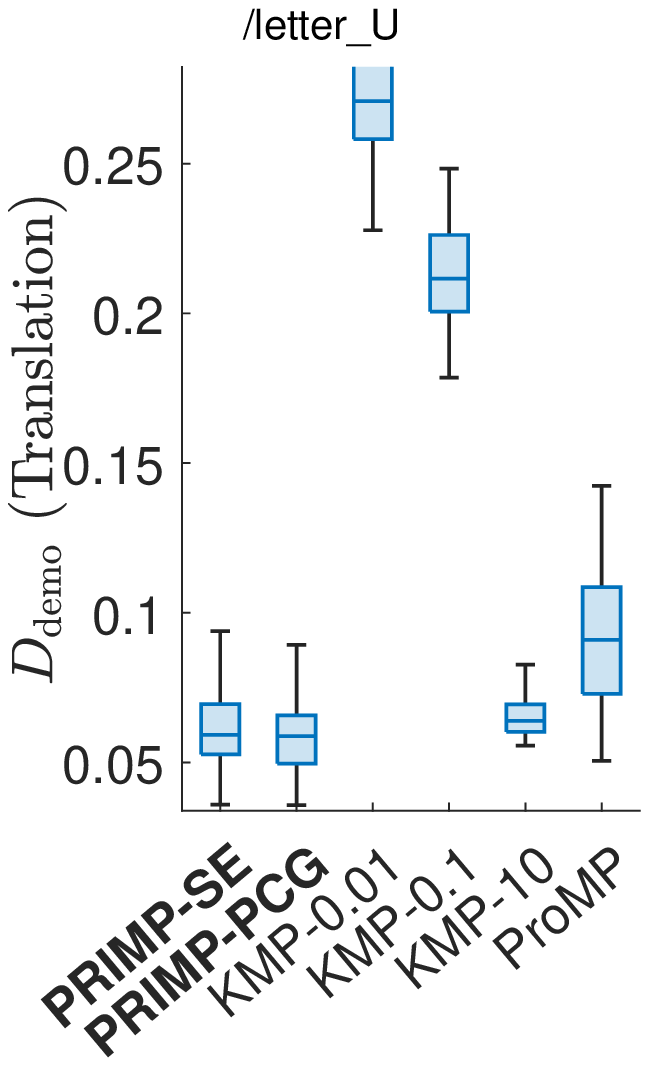} \label{fig:benchmark:lfd:simulation:distance:demo:letter_U}}
~
\subfloat[$D_{\rm demo}$, ``Letter S'']{\includegraphics[scale=0.4, trim={0 0 0 25}, clip]{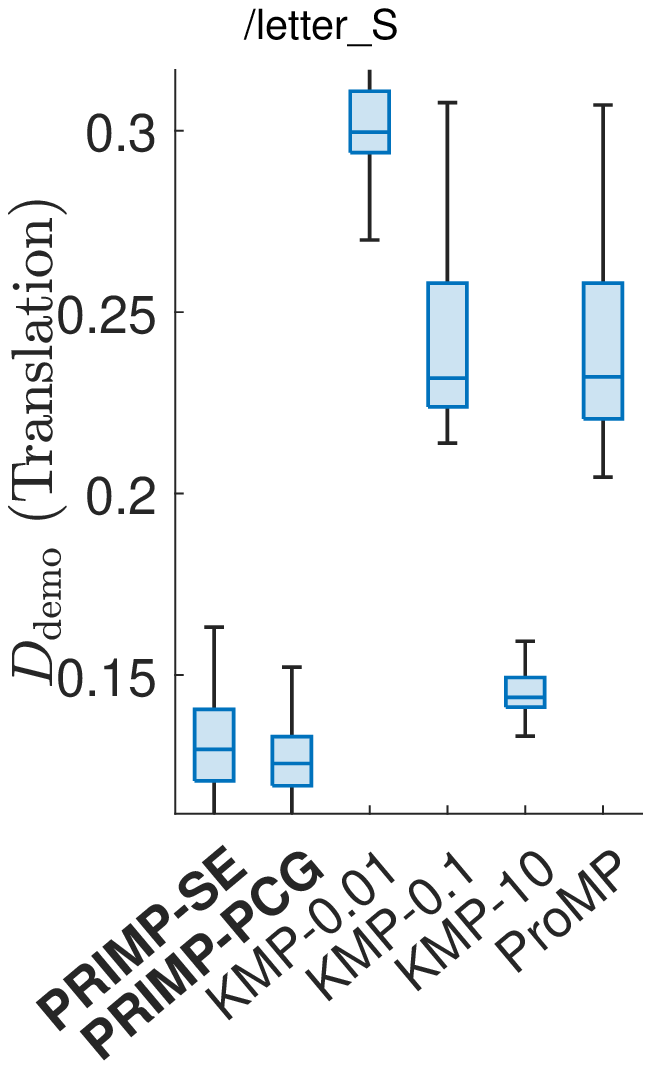} \label{fig:benchmark:lfd:simulation:distance:demo:letter_S}}
~
\subfloat[$D_{\rm via}$, ``Letter N'']{\includegraphics[scale=0.4, trim={0 0 0 25}, clip]{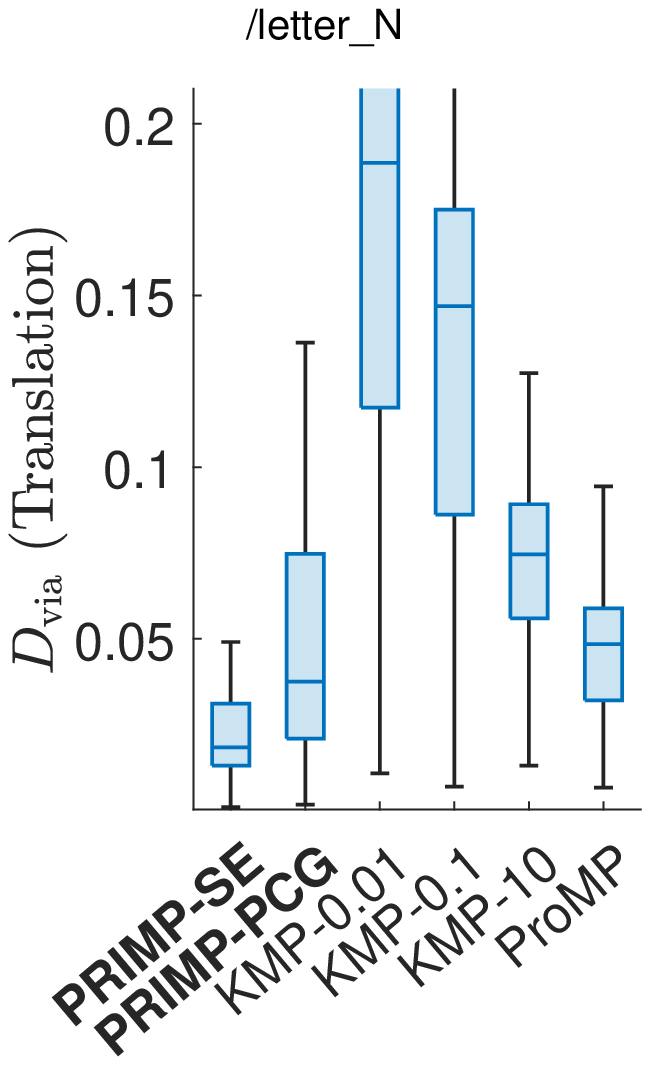} \label{fig:benchmark:lfd:simulation:distance:via:letter_N}}
~
\subfloat[$D_{\rm via}$, ``Letter U'']{\includegraphics[scale=0.4, trim={0 0 0 25}, clip]{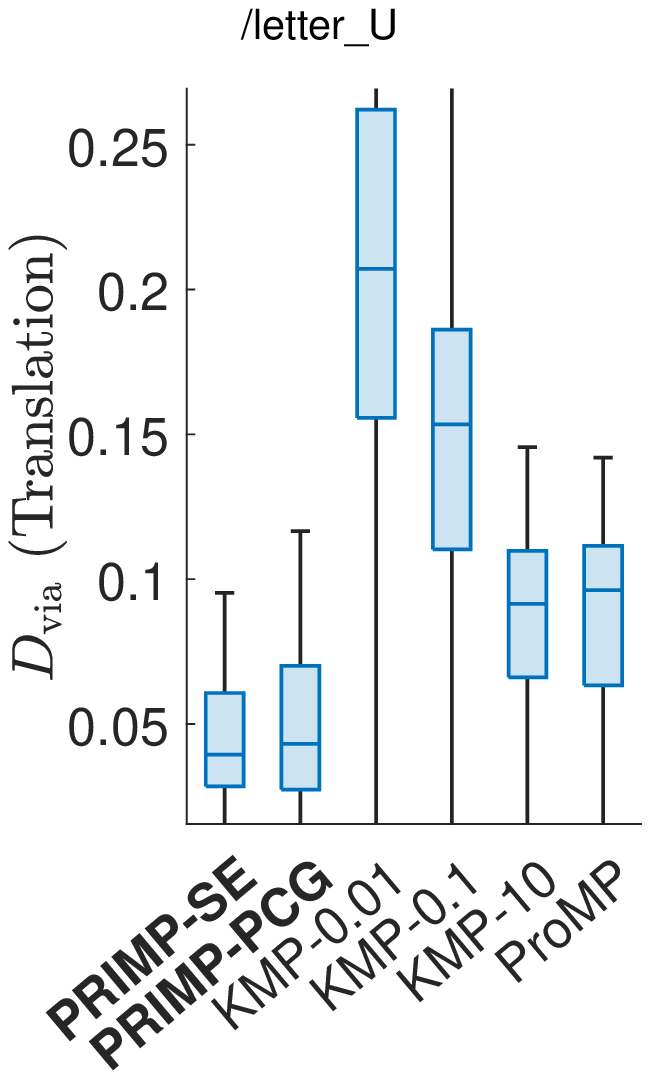} \label{fig:benchmark:lfd:simulation:distance:via:letter_U}}
~
\subfloat[$D_{\rm via}$, ``Letter S'']{\includegraphics[scale=0.4, trim={0 0 0 25}, clip]{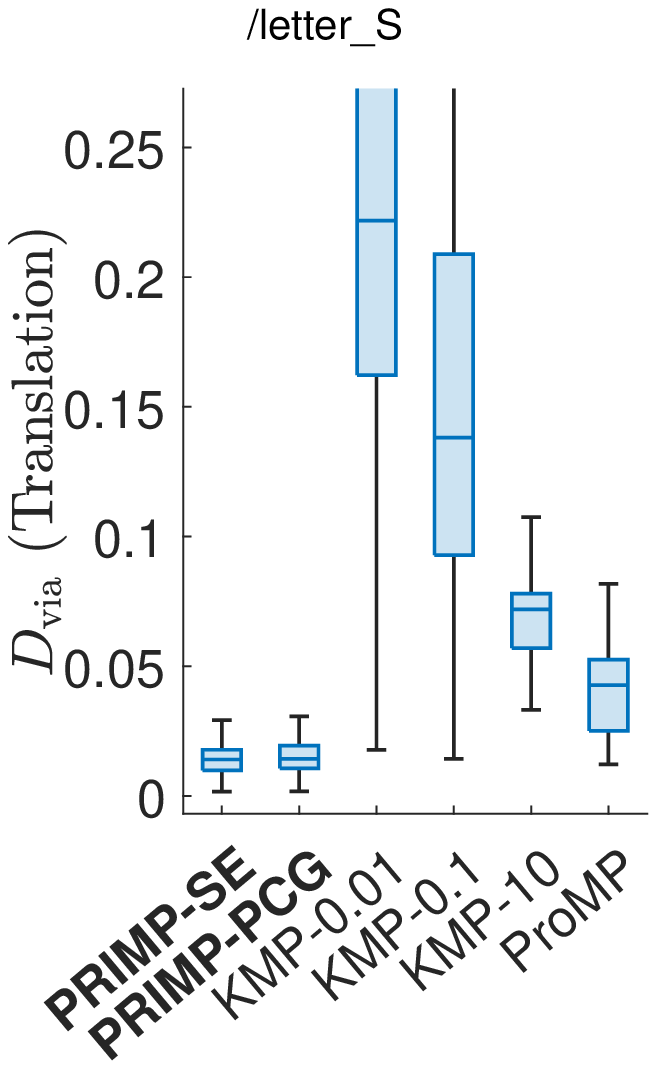} \label{fig:benchmark:lfd:simulation:distance:via:letter_S}}
\caption{Benchmark results for LfD methods in simulated data.}
\label{fig:benchmark:lfd:simulation:distance}
\end{figure*}

The simulations are conducted by virtually dragging the end effector to draw some symbols and letters in MoveIt. The orientation of the end effector is fixed. Figure \ref{fig:benchmark:lfd:simulation:distance} shows the benchmark results for the similarity to the demonstrated trajectories (Figs. \ref{fig:benchmark:lfd:simulation:distance:demo:letter_N}, \ref{fig:benchmark:lfd:simulation:distance:demo:letter_U}, \ref{fig:benchmark:lfd:simulation:distance:demo:letter_S}) and adaptability to via poses (Figs. \ref{fig:benchmark:lfd:simulation:distance:via:letter_N}, \ref{fig:benchmark:lfd:simulation:distance:via:letter_U}, \ref{fig:benchmark:lfd:simulation:distance:via:letter_S}).

\subsubsection{Physical demonstrations on real-world tasks} \label{sec:benchmark:lfd:real}
\begin{figure*}
\centering
\includegraphics[scale=0.52, trim={0 170 0 200}, clip]{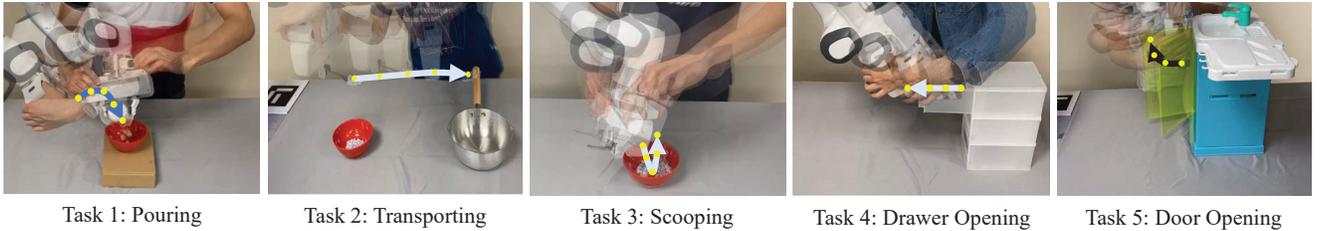}
\caption{Kinesthetic demonstrations for different tasks.}
\label{fig:benchmark:lfd:real:demo}
\end{figure*}

\begin{figure}
\centering
\subfloat[$D_{\rm demo}$, Task 1]{\includegraphics[scale=0.4, trim={0 0 0 25}, clip]{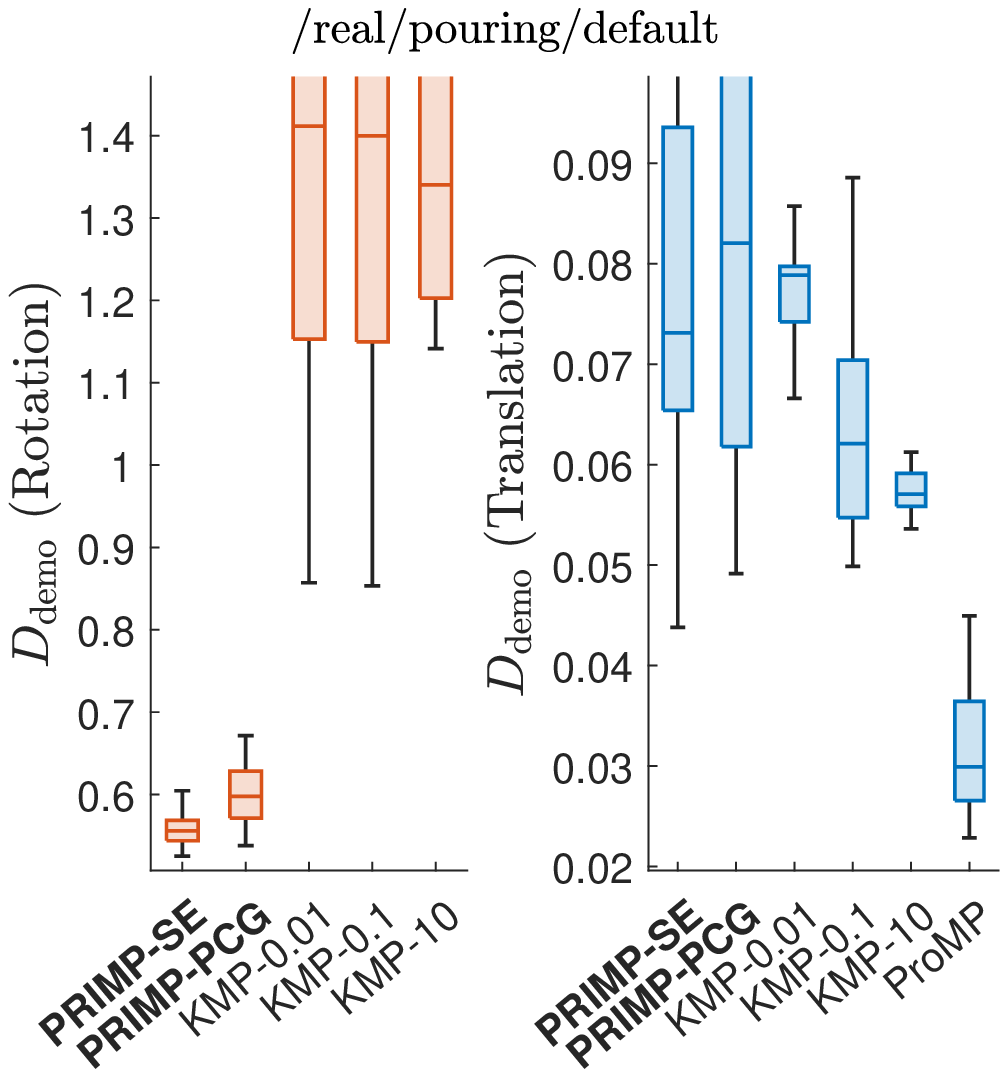} \label{fig:benchmark:lfd:real:distance:demo:cup_pouring}}
~
\subfloat[$D_{\rm via}$, Task 1]{\includegraphics[scale=0.4, trim={0 0 0 25}, clip]{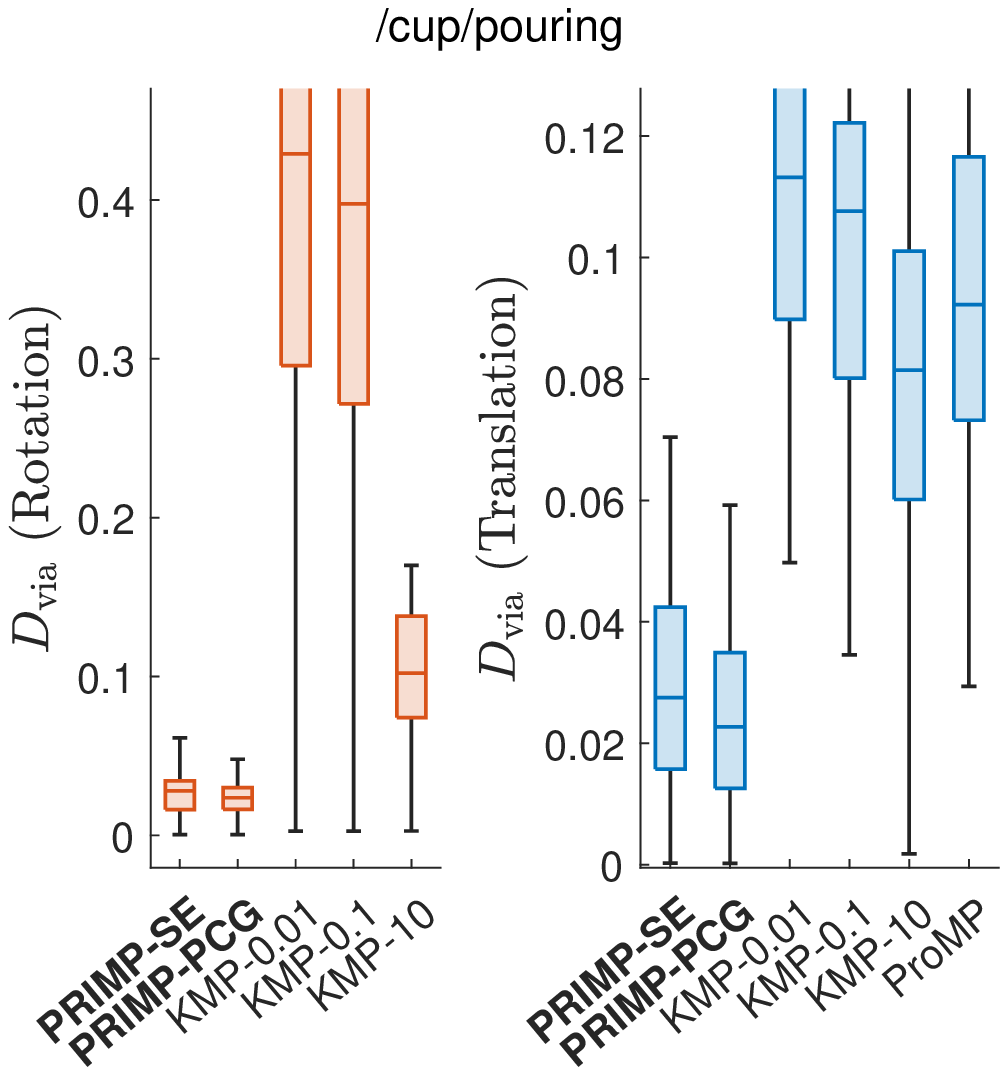} \label{fig:benchmark:lfd:real:distance:via:cup_pouring}}
\\
\subfloat[$D_{\rm demo}$, Task 4]{\includegraphics[scale=0.4, trim={0 0 0 25}, clip]{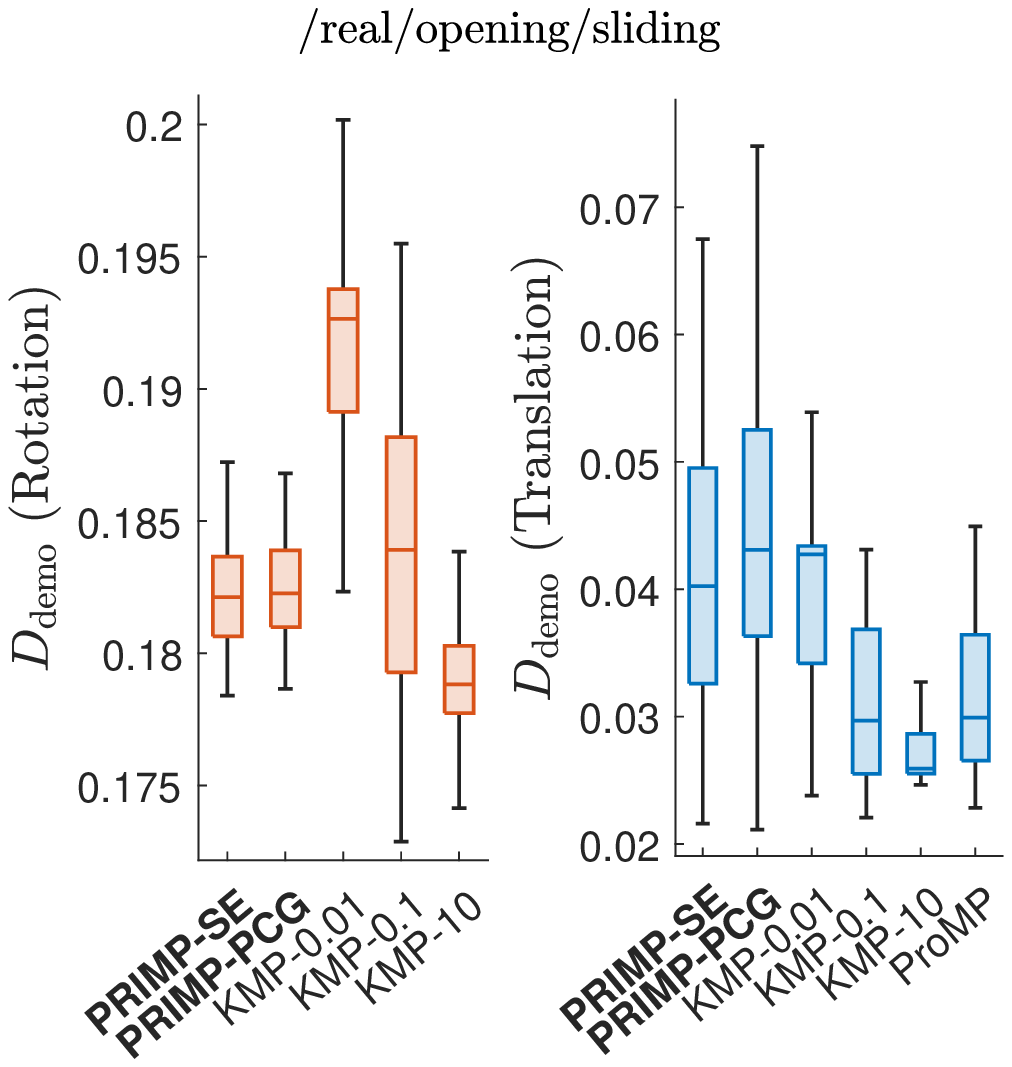} \label{fig:benchmark:lfd:real:distance:demo:drawer_opening}}
~
\subfloat[$D_{\rm via}$, Task 4]{\includegraphics[scale=0.4, trim={0 0 0 25}, clip]{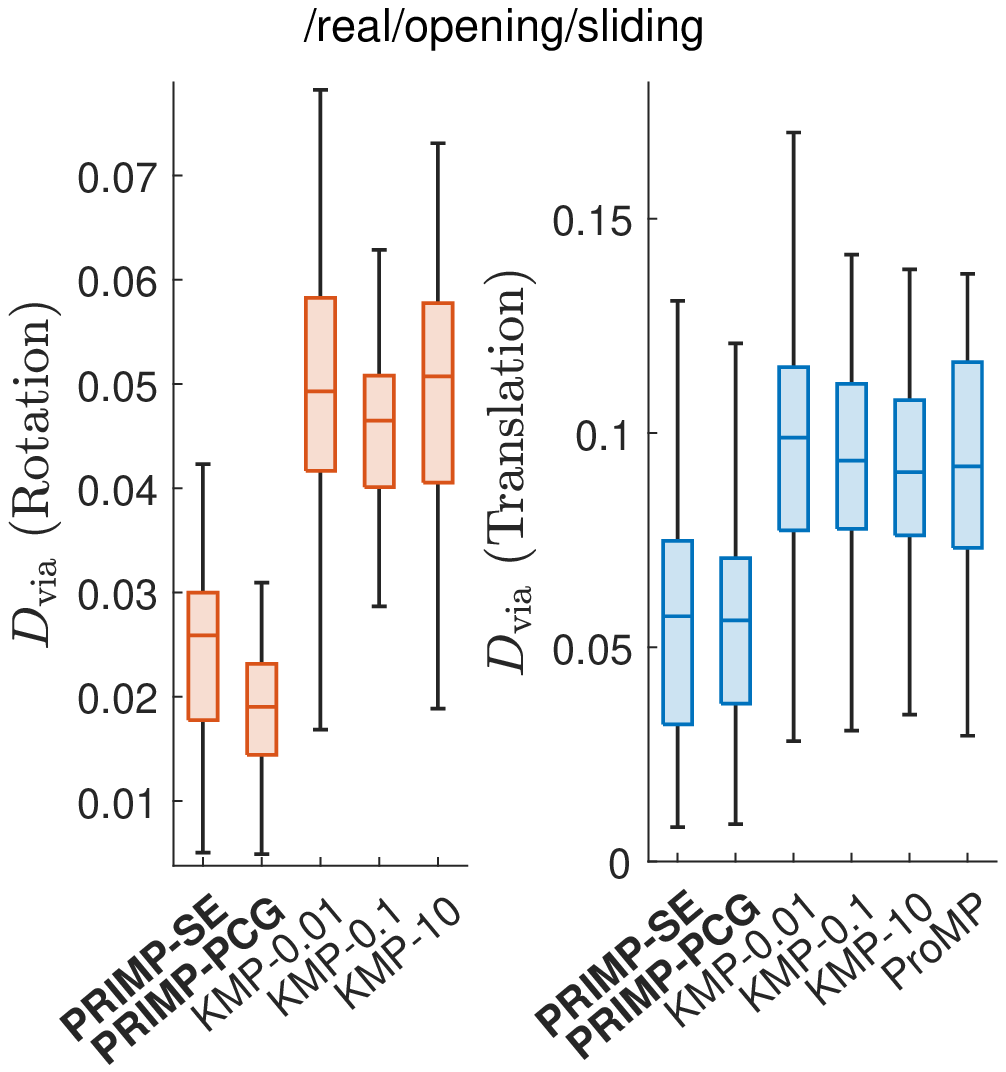} \label{fig:benchmark:lfd:real:distance:via:drawer_opening}}
\caption{Benchmark results for LfD methods in pouring (Task 1) and drawer opening (Task 4) tasks.}
\label{fig:benchmark:lfd:real:distance}
\end{figure}

We consider common tasks in daily household environments, which are summarized in Tab. \ref{tab:benchmark:tasks}. At first, for each task, the human demonstrator drags the end effector of the Panda robot, each with 5-10 trials (Fig. \ref{fig:benchmark:lfd:real:demo}). Then, with manually defined 50 different pairs of goals and via poses, different LfD algorithms are applied to re-produce the task. Similar distance metrics are used for evaluation, which are shown in Fig. \ref{fig:benchmark:lfd:real:distance}.

\begin{table*}[t]
\centering
\caption{Daily objects and common tasks used in the benchmark.}
\begin{tabular}{ccccc}
\toprule
Task ID & Object & Tool & Motion primitives & Task description \\
\midrule
1 & Container & Cup/spoon & Pouring & Pour particles to a container \\
2 & -- & Spoon & Transporting & Transport particles without spillage \\
3 & Container & Spoon & Scooping & Scoop particles from a container \\
4 & Drawer & Handle & Opening & Open the sliding door of a drawer \\
5 & Door & Handle & Opening & Open the door with hinge \\
\bottomrule
\end{tabular}
\label{tab:benchmark:tasks}
\end{table*}

\subsection{Properties of PRIMP} \label{sec:benchmark:PRIMP_property}

\subsubsection{Extrapolation}
\begin{figure}[t]
\centering
\subfloat[PRIMP]{\includegraphics[scale=0.35, trim=50 0 20 0, clip]{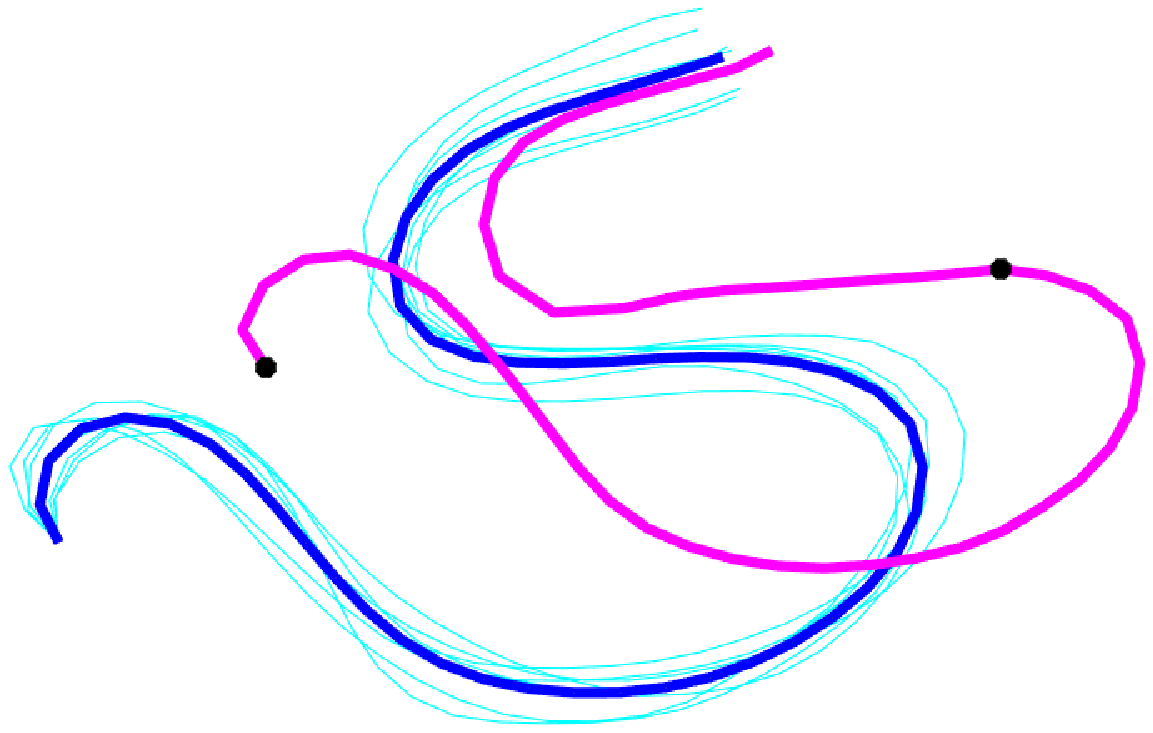}}
~
\subfloat[Orientation-KMP]{\includegraphics[scale=0.35, trim=50 0 20 0, clip]{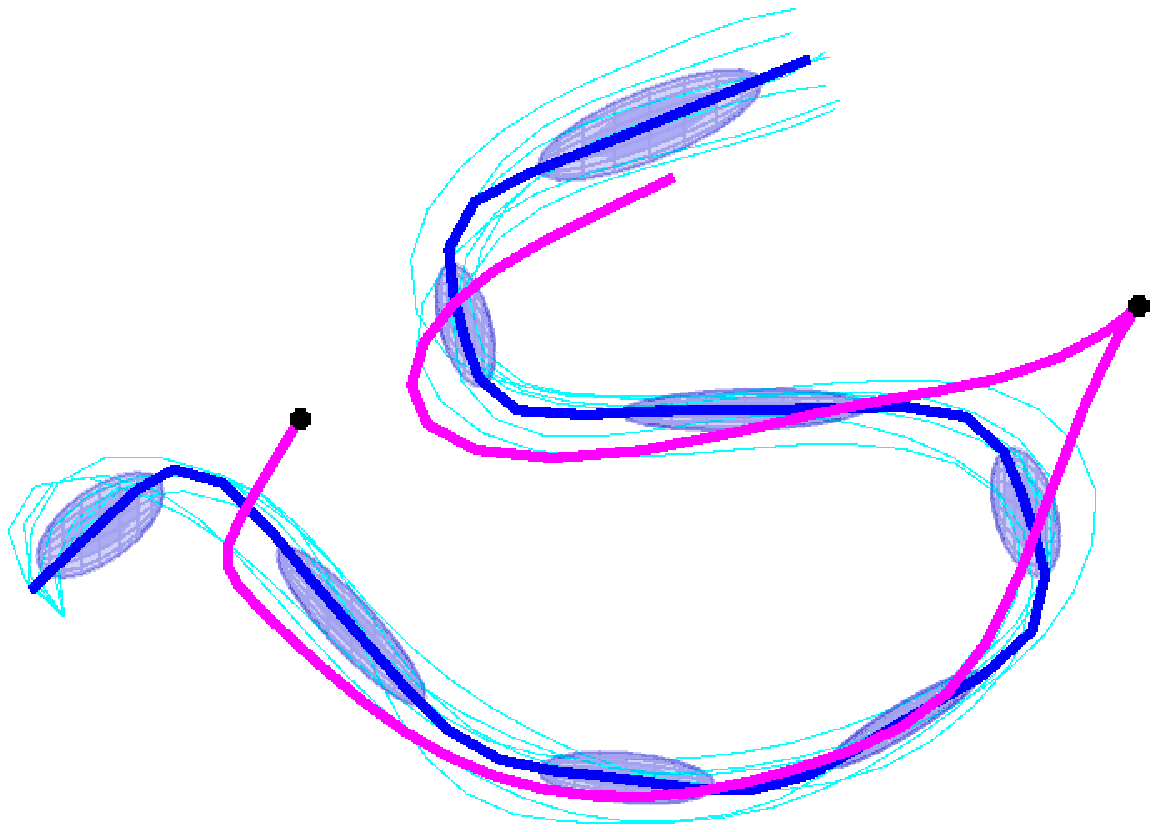}}
\caption{Qualitative comparisons for extrapolation between PRIMP and Orientation-KMP. Positions of via poses are indicated by black dots. Solid thick blue and magenta curves are the means of the demonstrated and reproduced trajectories, respectively. For Orientation-KMP (b), blue ellipsoids indicate the level surface of the Gaussian mixture model.}
\label{fig:benchmark:PRIMP:extrapolation}
\end{figure}

One of the most desirable properties of an LfD method is the ability to adapt to extrapolations when via poses are out of the distributions of the demonstrations. A qualitative comparison is shown for the case of extrapolation between PRIMP and Orientation-KMP methods in Fig. \ref{fig:benchmark:PRIMP:extrapolation}.

\subsubsection{Learning from a single demonstration}
\begin{figure}[t]
\centering
\subfloat[Task 1]{\includegraphics[scale=0.35, trim={50 0 100 20}, clip]{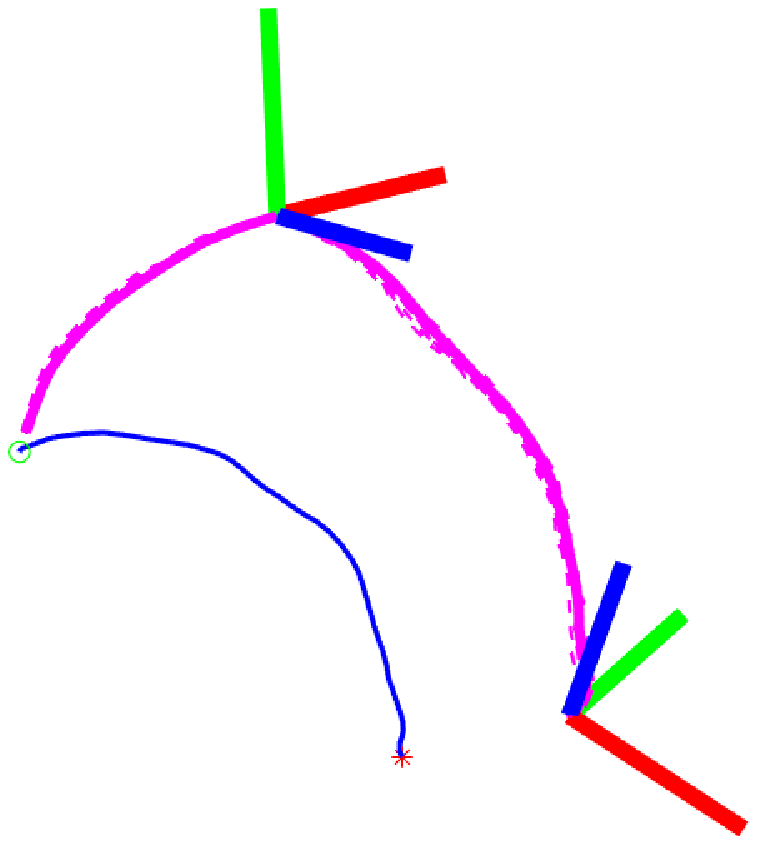}}
~
\subfloat[Task 4]{\includegraphics[scale=0.35, trim={50 0 80 20}, clip]{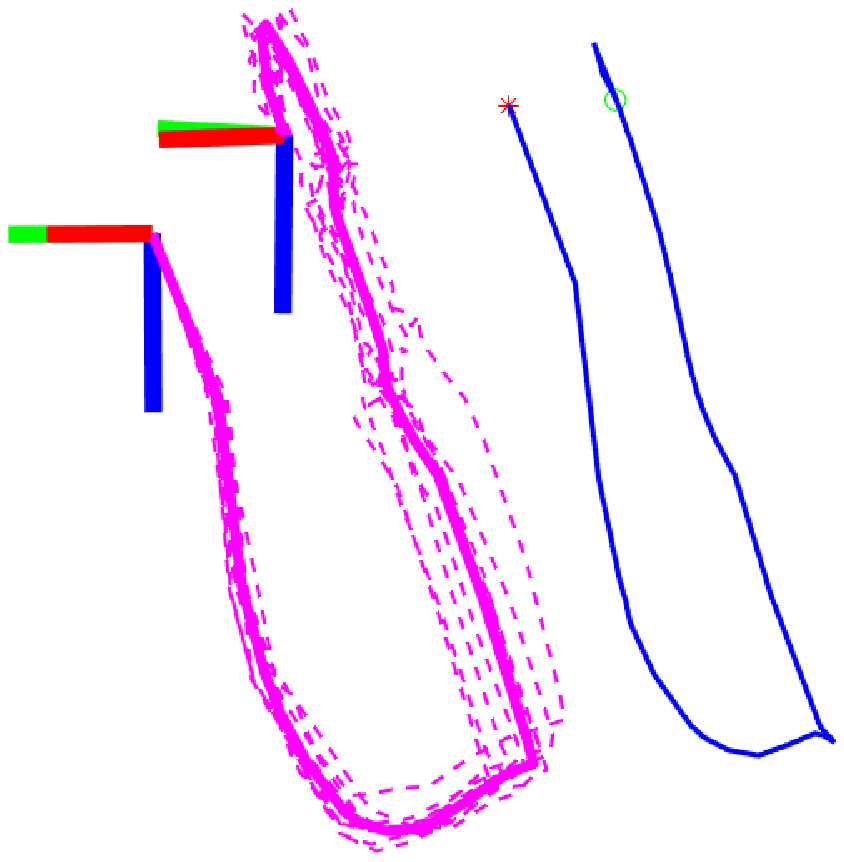}}
\caption{PRIMP learning from a single demonstration. The Blue curve is the only demonstrated trajectory. Solid and dashed magenta curves are the mean and samples from the adapted trajectory after conditioning, respectively.}
\label{fig:benchmark:PRIMP:single_demo}
\end{figure}

The ability of PRIMP to learn from a single demonstration is illustrated in Fig. \ref{fig:benchmark:PRIMP:single_demo} qualitatively. The real-world tasks are selected, each of which only consists of one demonstrated trajectory.

\subsection{Comparisons on guided motion planning} \label{sec:benchmark:planner}
\begin{figure}
\centering
\subfloat[Sparse]{\includegraphics[scale=0.08]{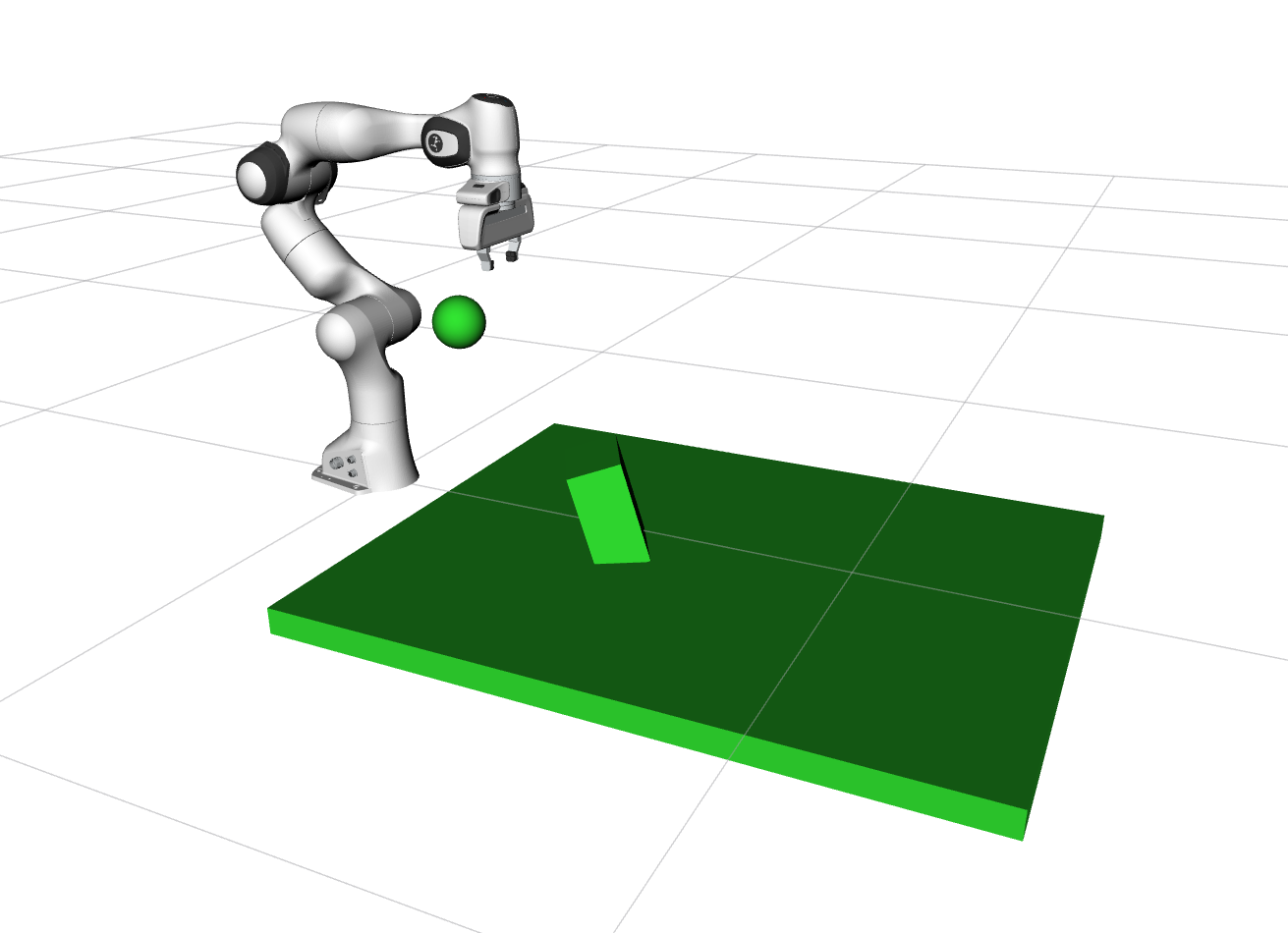}}
~
\subfloat[Cluttered]{\includegraphics[scale=0.08]{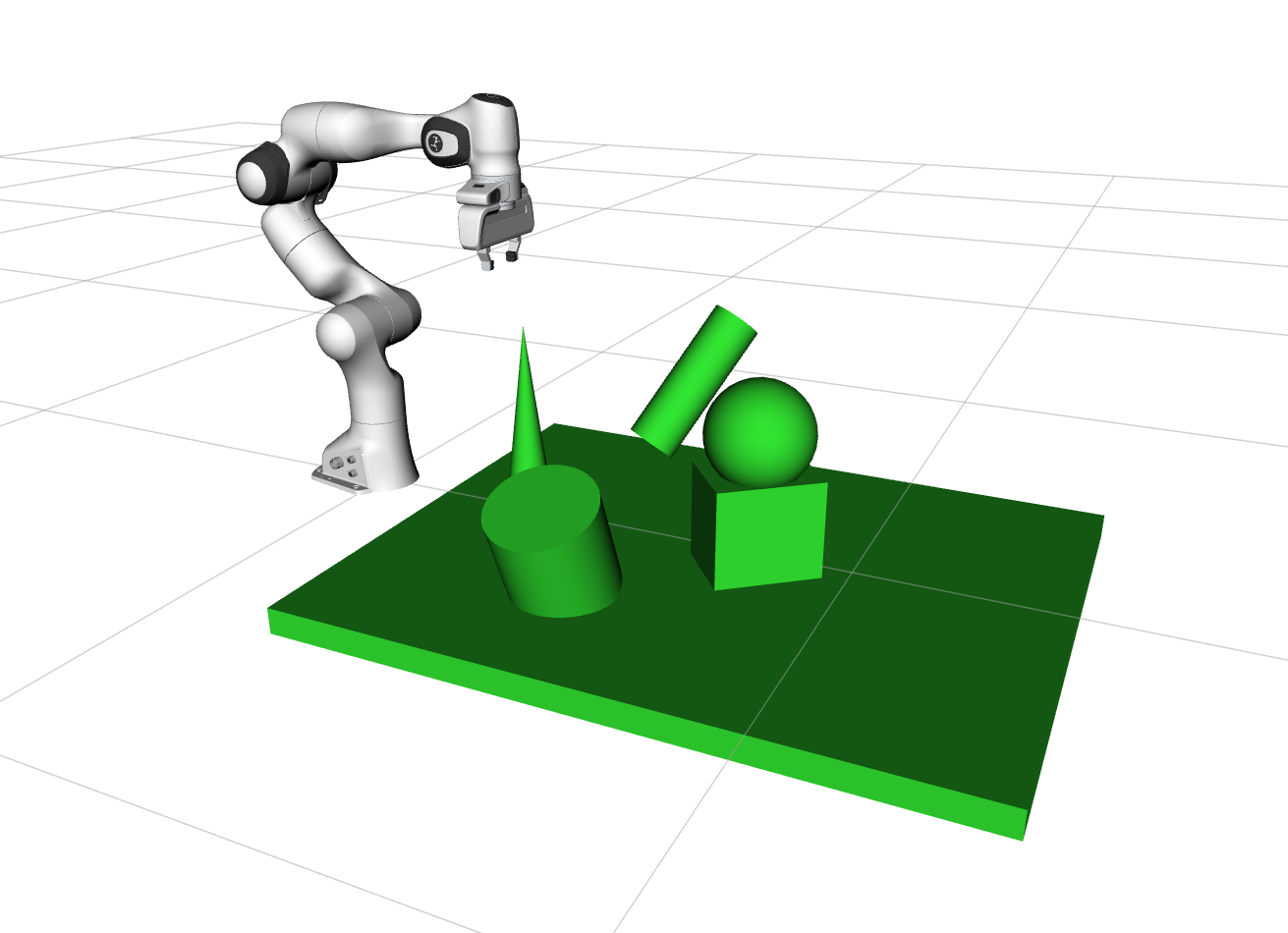}}
~
\subfloat[Narrow]{\includegraphics[scale=0.08]{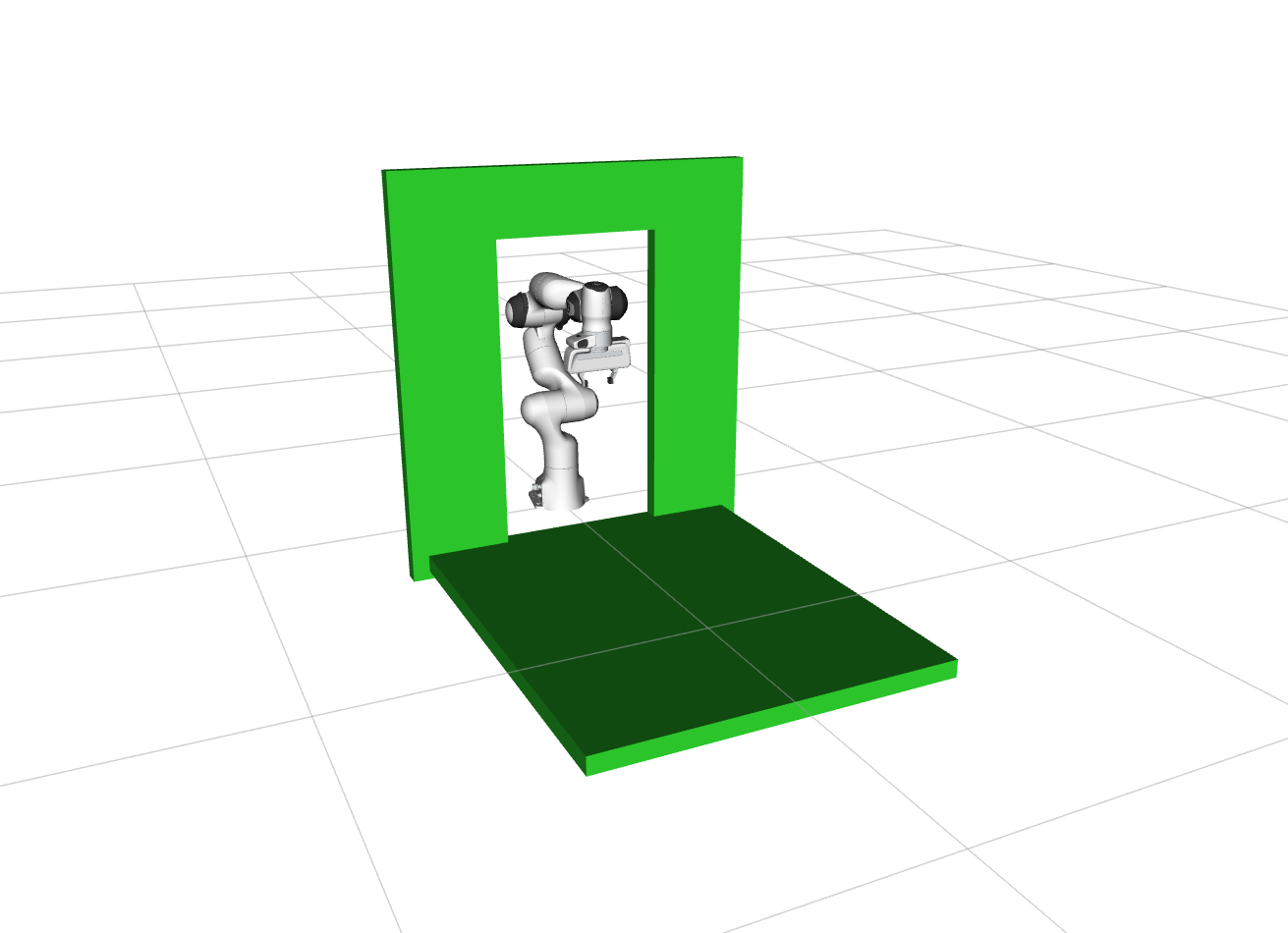}}
\caption{Simulated planning scenes for planning benchmarks.}
\label{fig:benchmark:planning:scene}
\end{figure}

\begin{figure}
\centering
\subfloat[Task 1, empty scene]{\includegraphics[scale=0.4, trim={0 0 0 20}, clip]{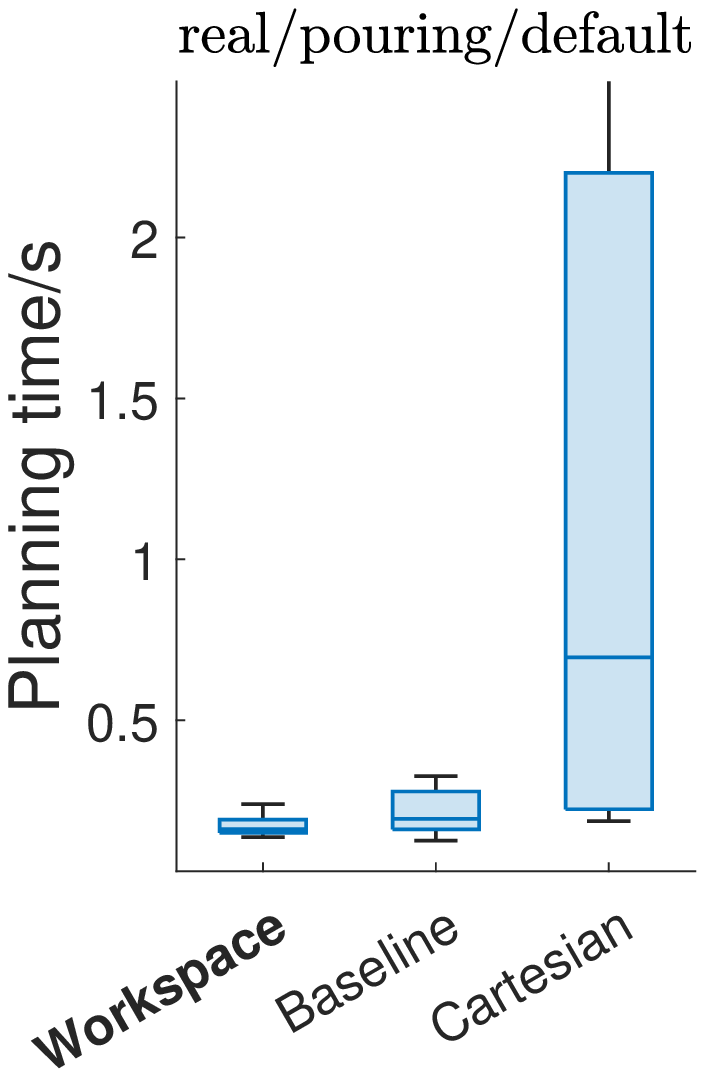} \label{fig:benchmark:planning:time:empty:cup_pouring}}
~
\subfloat[Task 1, narrow scene]{\includegraphics[scale=0.4, trim={0 0 0 20}, clip]{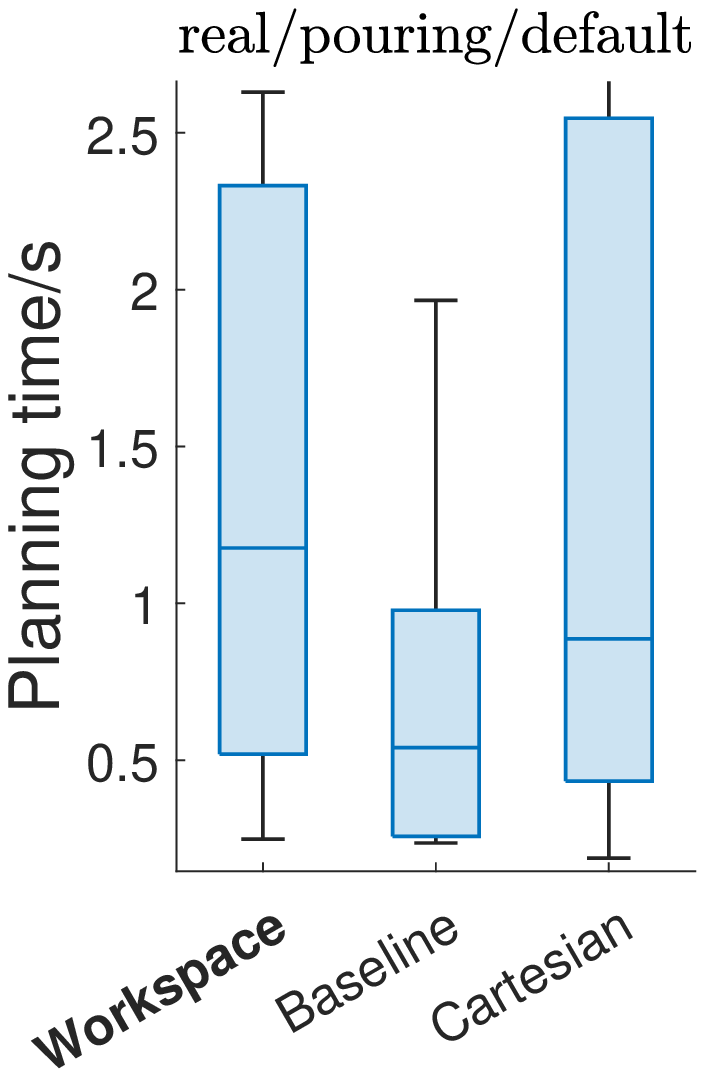} \label{fig:benchmark:planning:time:narrow:cup_pouring}}
\\
\subfloat[Task 4, sparse scene]{\includegraphics[scale=0.4, trim={0 0 0 20}, clip]{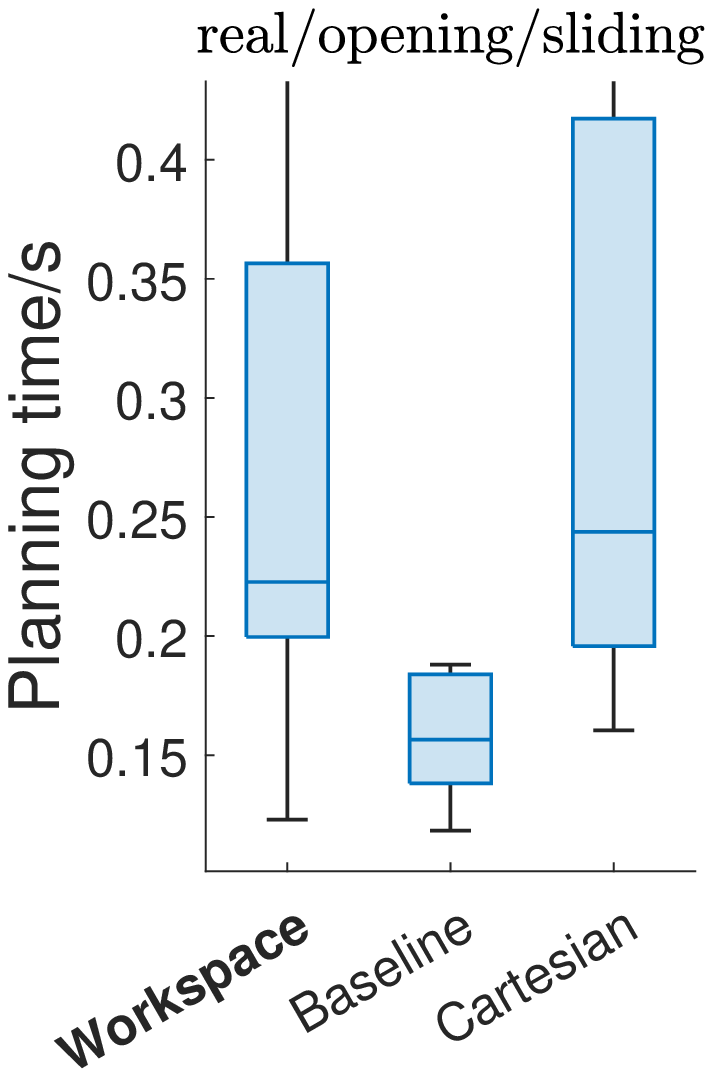} \label{fig:benchmark:planning:time:sparse:drawer_opening}}
~
\subfloat[Task 4, cluttered scene]{\includegraphics[scale=0.4, trim={0 0 0 20}, clip]{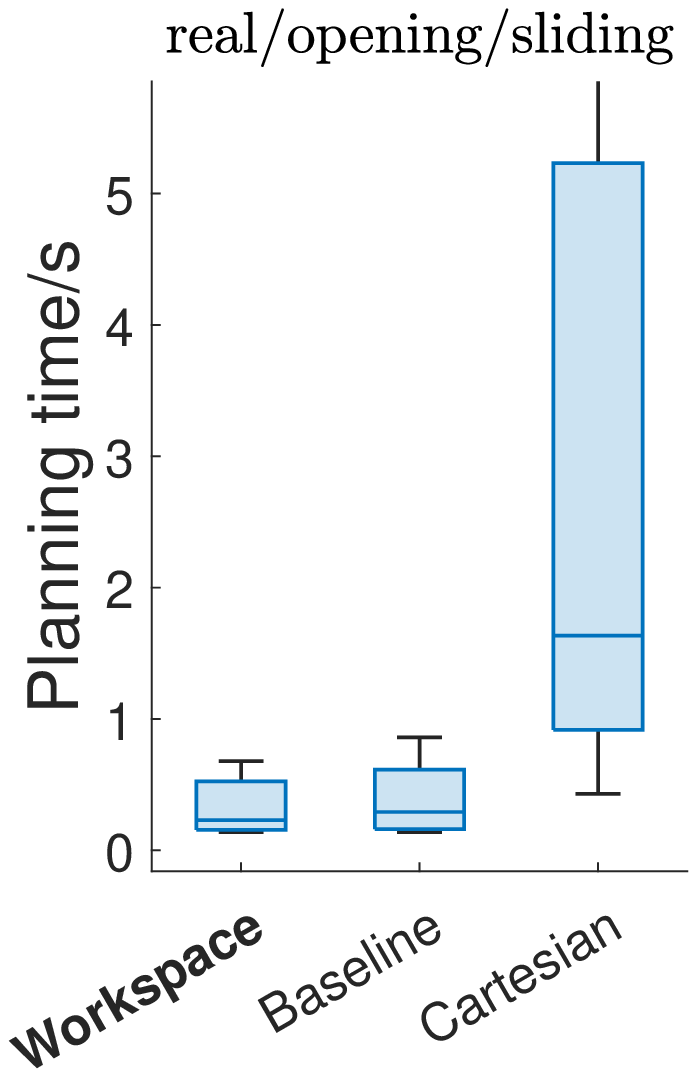} \label{fig:benchmark:planning:time:cluttered:drawer_opening}}
\caption{Planning time comparisons for different STOMP variants in pouring (Task 1) and drawer opening (Task 4) tasks.}
\label{fig:benchmark:planning:time}
\end{figure}

\begin{figure}
\centering
\subfloat[Task 1, empty scene]{\includegraphics[scale=0.4, trim={0 0 0 25}, clip]{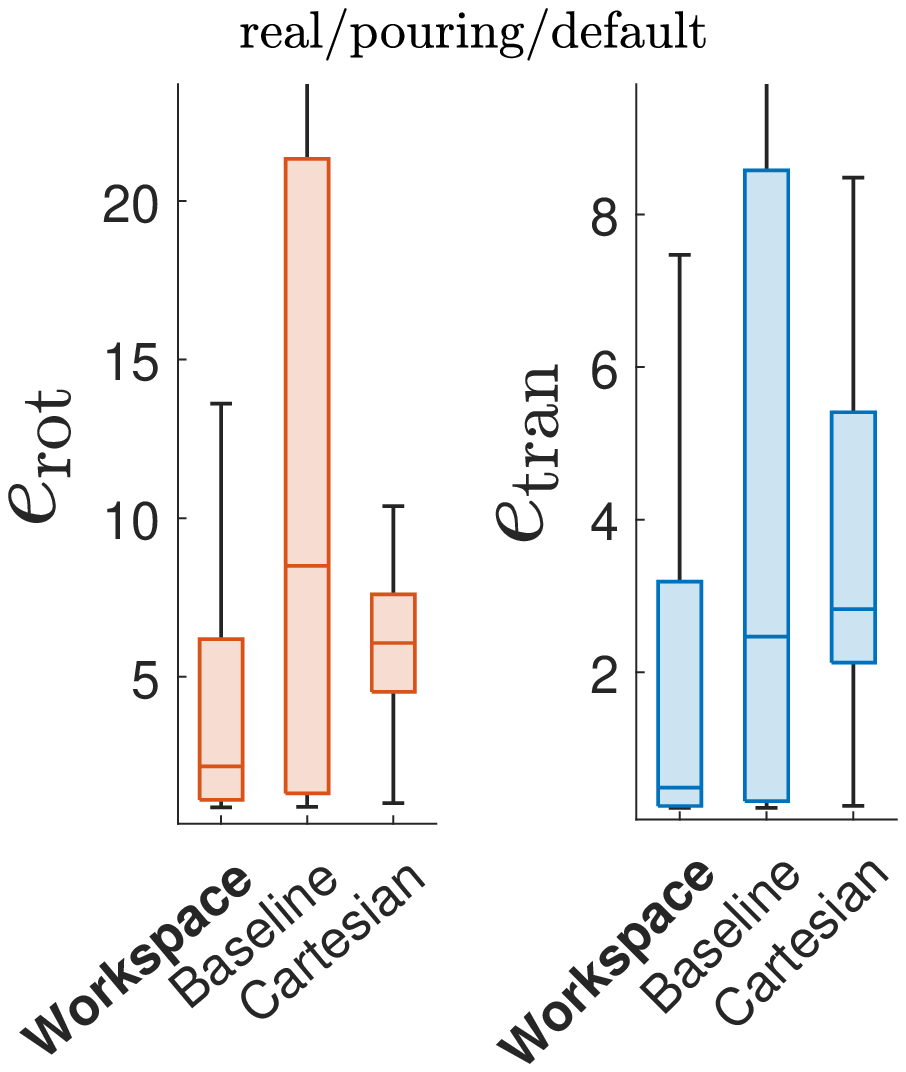} \label{fig:benchmark:planning:error:empty:cup_pouring}}
~
\subfloat[Task 1, narrow scene]{\includegraphics[scale=0.4, trim={0 0 0 25}, clip]{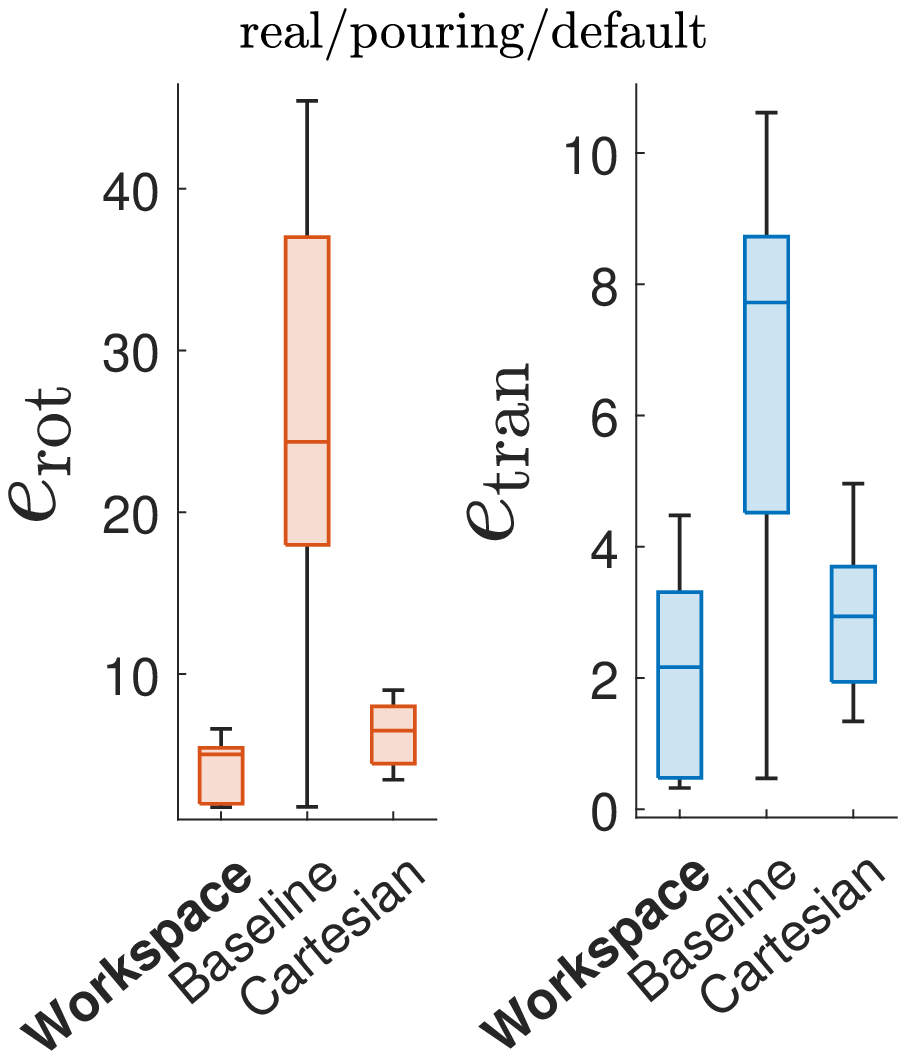} \label{fig:benchmark:planning:error:narrow:cup_pouring}}
\\
\subfloat[Task 4, sparse scene]{\includegraphics[scale=0.4, trim={0 0 0 25}, clip]{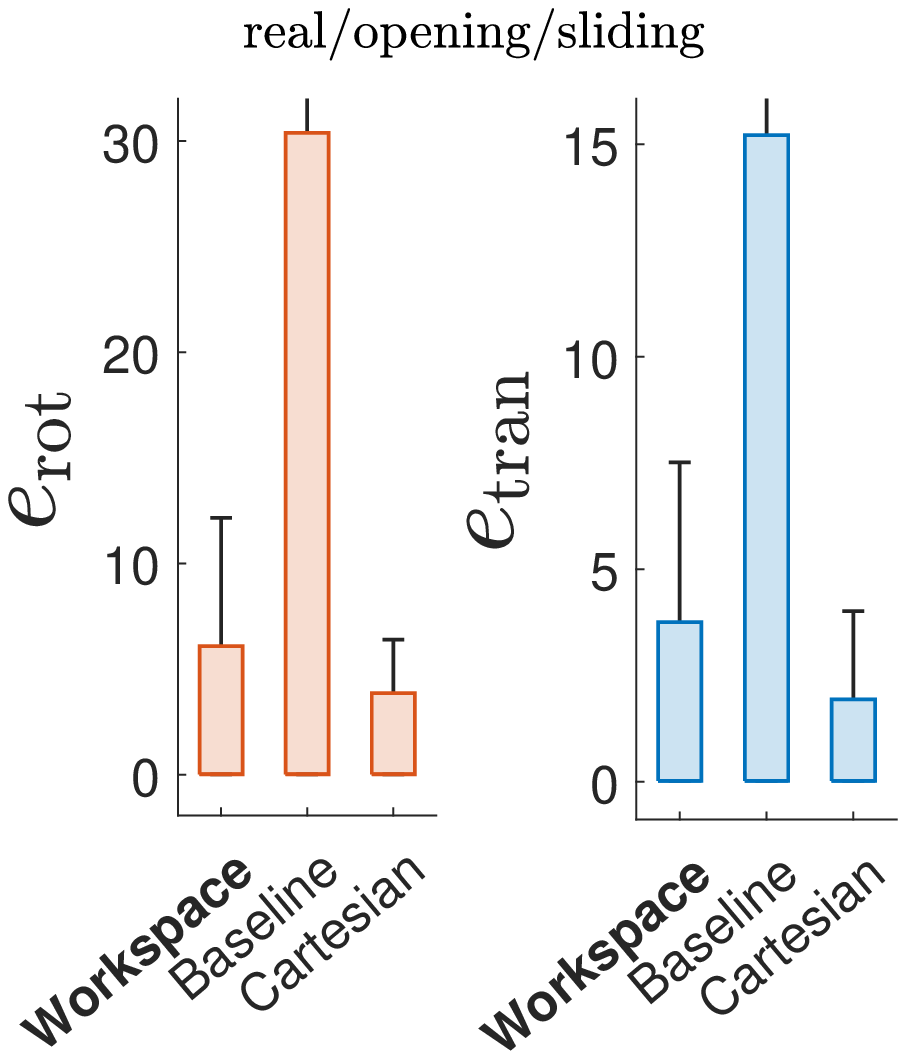} \label{fig:benchmark:planning:error:sparse:drawer_opening}}
~
\subfloat[Task 4, cluttered scene]{\includegraphics[scale=0.4, trim={0 0 0 25}, clip]{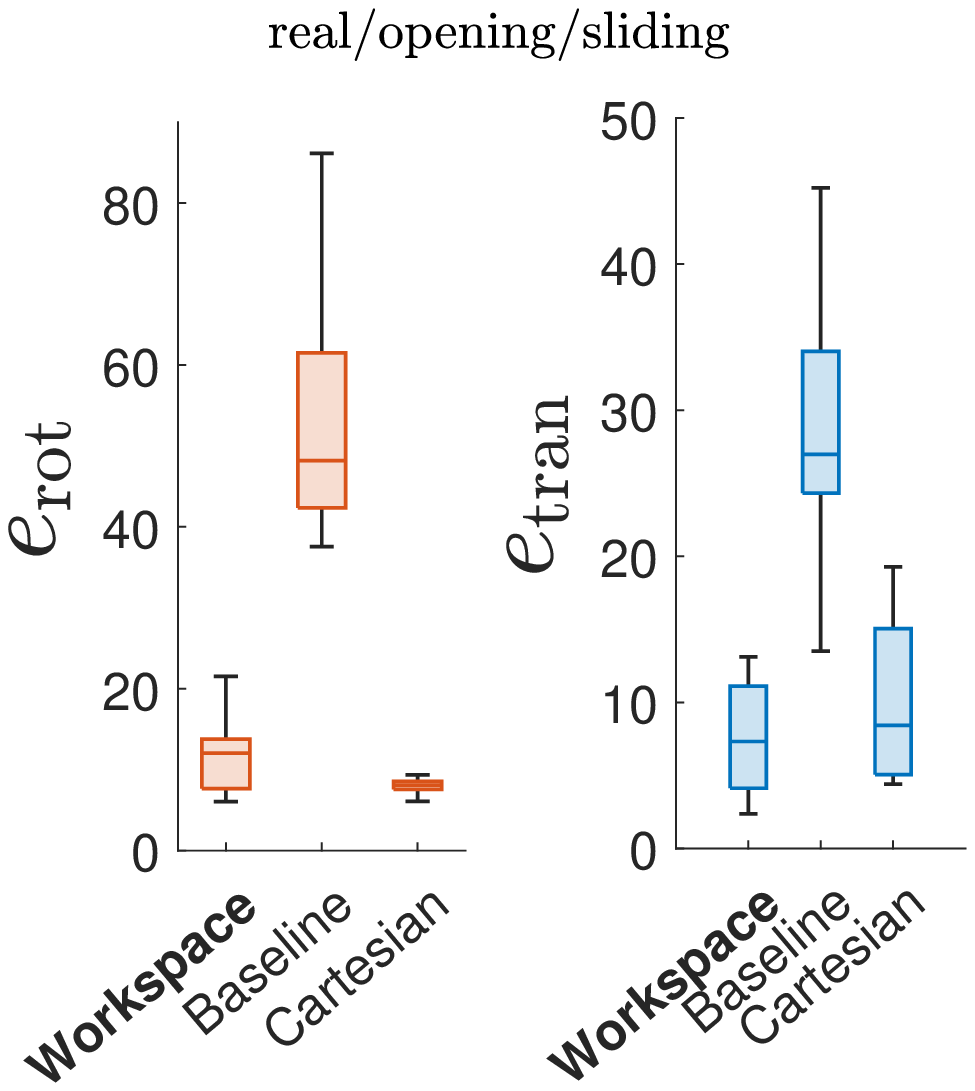} \label{fig:benchmark:planning:error:cluttered:drawer_opening}}
\caption{Comparisons of distance between planned trajectory and reference trajectory for different variants of STOMP in pouring (Task 1) and drawer opening (Task 4) tasks. The metric is computed using Eq. \eqref{eq:benchmark:e_ref}.}
\label{fig:benchmark:planning:error}
\end{figure}

Benchmarks for motion planning are conducted, which include manually defined environments using simple geometric primitives in simulation (as in Fig. \ref{fig:benchmark:planning:scene}). An empty scene is also added for evaluation. We compare the proposed Workspace-STOMP planner with: (1) vanilla STOMP \cite{kalakrishnan2011stomp}; and (2) Cartesian-guided STOMP \cite{dobivs2021cartesian}. The real-world tasks, as defined in Sec. \ref{sec:benchmark:lfd:real}, are applied here. The initial trajectories for all the planners are set to be the mean of the learned distribution from PRIMP and converted to joint space using inverse kinematics. The benchmark results for different tasks in different planning scenes include planning time (Fig. \ref{fig:benchmark:planning:time}), success rate (Tab. \ref{tab:benchmark:planning:sr}) and distance between planned trajectory and mean of the reference trajectory distribution (Fig. \ref{fig:benchmark:planning:error}).

\begin{table}[t]
\centering
\caption{Comparisons for success rate among different planners (the cases when all planners failed are omitted).}
\begin{tabular}{ccccc}
\toprule
Scene & Task & STOMP & Cartesian-STOMP & {\bf Workspace-STOMP} \\
\midrule
\multirow{4}{*}{Empty} & 1 & $100\%$ & $85\%$ & ${\bf 100\%}$ \\
 & 2 & $100\%$ & $40\%$ & ${\bf 100\%}$ \\
 & 3 & $100\%$ & $100\%$ & ${\bf 100\%}$ \\
 & 4 & $100\%$ & $100\%$ & ${\bf 100\%}$ \\
\midrule
\multirow{4}{*}{Sparse} & 1 & ${\bf 12\%}$ & $10\%$ & $10\%$ \\
 & 2 & ${\bf 8\%}$ & $4\%$ & $0\%$ \\
 & 3 & $6\%$ & $4\%$ & ${\bf 6\%}$ \\
 & 4 & $62\%$ & $62\%$ & ${\bf 62\%}$ \\
\midrule
\multirow{2}{*}{Cluttered} & 3 & $66\%$ & $64\%$ & ${\bf 66\%}$ \\
 & 4 & $30\%$ & $28\%$ & ${\bf 30\%}$ \\
\midrule
\multirow{3}{*}{Narrow} & 1 & ${\bf 26\%}$ & $14\%$ & $14\%$ \\
 & 2 & $8\%$ & $12\%$ & ${\bf 32\%}$ \\
 & 3 & ${\bf 22\%}$ & $0\%$ & $4\%$ \\
\bottomrule
\end{tabular}
\label{tab:benchmark:planning:sr}
\end{table}

\section{Physical experiments} \label{sec:physical-experiments}
A robotic system is proposed, which includes PRIMP, Workspace-STOMP and robot imagination \cite{wu2020can}. The tasks and the corresponding motion primitives are the same in Tab. \ref{tab:benchmark:tasks}. Physical experiments using the Franka Emika Panda robot are conducted.

\subsection{Workflow}
\begin{figure*}
\centering
\includegraphics[scale=0.52, trim = 10 30 10 30, clip]{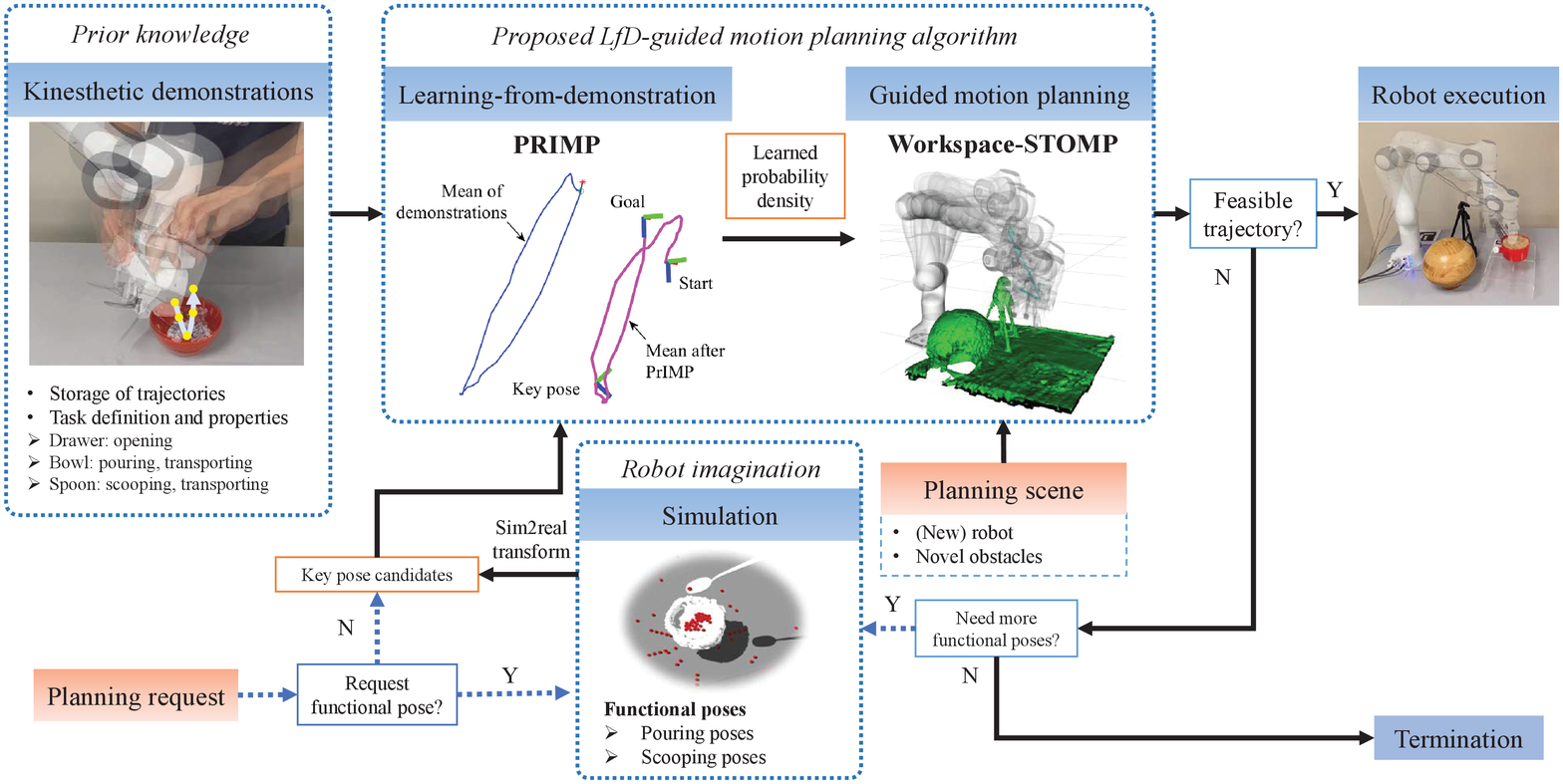}
\caption{Workflow of the proposed robotic system including PRIMP, Workspace-STOMP and robot imagination.}
\label{fig:experiments:workflow}
\end{figure*}

The general workflow of the proposed robotic system is shown in Fig. \ref{fig:experiments:workflow}. For each motion primitive, human operators firstly conduct several demonstrations by dragging the robot end effector to fulfill the specific task. The trajectory of the end effector poses for each demonstration is recorded. For a new planning request, \textit{key poses} for the robot are generated from manual inputs, ArUco tags \cite{garrido2014automatic} or robot imagination module (as in Sec. \ref{sec:experiments:imagination}). A set of key pose candidates are then fed into PRIMP to condition the trajectory probabilistic distribution. The learned distribution is then used to guide the STOMP planner with new planning scenes, which include novel obstacles. Once a feasible trajectory is found by Workspace-STOMP, the robot executes the planned motion to fulfill the designated task. If there is no feasible trajectory, more key pose candidates are generated for re-planning.



\subsection{3D reconstruction of the planning scene}
A new scenario is obtained by 3D reconstructions using an RGB+D camera. It is mounted at the robot end effector and moved to several pre-defined configurations (the details are referred to \cite{wu2020can}). The object of interest, \ie container or drawer, is segmented from the planning scene by placing it within a transparent table whose size is pre-defined and pose is indicated by an ArUco tag. After the scanning, the object of interest and obstacles outside the transparent table are constructed as mesh models separately. The object of interest is then loaded into the robot imagination module for tasks like pouring and scooping. In the planning module, only the novel obstacles are loaded into the planning scene.

\subsection{Prior knowledge for tasks and tools} \label{sec:experiments:prior_knowledge}
Prior knowledge of each task includes the method to obtain \textit{functional poses}, \textit{key poses} and transformation from functional pose to key pose. The functional pose is obtained by (1) robot imagination, such as the pouring pose and scooping pose; or (2) the tag that indicates the start/goal pose for transporting or opening. The key pose is defined with respect to the specific link of the robot arm that is the same as the recorded one during the demonstrations. One key pose is the start pose, which is pre-defined for each task. Other key poses are generated depending on whether robot imagination is applied. For the tasks that require robot imagination, \ie pouring and scooping, only the time parameter is defined a priori. But for tasks like opening, the goal pose is computed as prior knowledge by considering the dimension of the moving part and the mode of movement. The prior knowledge for relative transformation from the functional pose to the key pose is defined for each tool, \ie through the geometries of the handle, spoon and cup and the corresponding grasping point for the robot gripper. Table \ref{tab:experiments:prior_knowledge} summarizes the prior knowledge for each task, including the process that generates key poses and the corresponding time parameter (\ie $t \in [0,1]$).

\begin{table}[t!]
\centering
\caption{Prior knowledge for different tasks.}
\begin{tabular}{cccc}
\toprule
Task ID & Key pose & $t \in [0,1]$ & Generation method \\
\midrule
\multirow{2}{*}{1} & Start & 0.0 & Certain distance above object \\
& Pouring & 1.0 & Pouring imagination \\
\midrule
\multirow{2}{*}{2} & Start & 0.0 & ArUco tag \\
& Goal & 1.0 & ArUco tag \\
\midrule
\multirow{3}{*}{3} & Start & 0.0 & Certain distance above object \\
& Scooping & 0.5 & Scooping imagination \\
& Goal & 1.0 & Certain distance above object \\
\midrule
\multirow{2}{*}{4}  & Start & 0.0 & ArUco tag \\
& Goal & 1.0 & Computed from object geometry \\
\midrule
\multirow{2}{*}{5}  & Start & 0.0 & ArUco tag \\
& Goal & 1.0 & Computed from object geometry \\
\bottomrule
\end{tabular}
\label{tab:experiments:prior_knowledge}
\end{table}

\subsection{Robot imagination} \label{sec:experiments:imagination}
The robot imagination provides functional poses for the pouring and scooping tasks, which are performed using Gazebo physics simulator. The pouring imagination applies the same principle with \cite{wu2020can}, while the scooping imagination is novel in this work. The process for simulating the scooping tasks is described as follows.

\begin{figure}
\centering
\includegraphics[scale=0.55, trim = 260 200 260 200, clip]{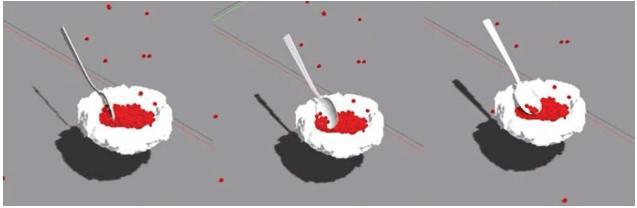} 
\caption{Robot imagination process for scooping. The container is represented by the white meshed object.}
\label{fig:experiments:imagination}
\end{figure}

The tool for scooping is a common spoon model, whose center is defined as the center of the rim of its head part. After the object of interest is loaded into the simulator, particles are dropped above the object's bounding box. Once there are some particles contained inside, the one with the minimum height is recorded, denoted as the \textit{bottom particle}. Its position is treated as the potential scooping position. Then, the spoon is placed on top of this bottom particle with some random variations in $x-y$ direction. The orientation is sampled as a uniformly distributed random rotation around the global $z$-axis. Before scooping, the spoon is rotated for $90^{\circ}$ around its longer principle axis so that the head is facing the side. Then, the spoon is slowly rotated around the longer principle axis until the head is facing the top (as shown in Fig. \ref{fig:experiments:imagination}). The current pose is a candidate of the \textit{functional (scooping) pose}. Then, the spoon is lifted above the object to count the particles. Once there is any particle within the spoon head, the functional pose candidate is recorded.

The obtained functional poses are transformed into key poses for the robot by prior knowledge defined in Sec. \ref{sec:experiments:prior_knowledge} and will be used in PRIMP. The number of attempts ($N_{\rm attempt}$) and the maximum number of functional poses ($N_{\rm pose}$) are defined by users. The imagination process is considered a success if $N_{\rm pose} > 0$ after $N_{\rm attempt}$ trials, and failure otherwise.


\subsection{Results}
\begin{figure*}
\centering
\includegraphics[scale=0.52, trim = 0 180 0 180, clip]{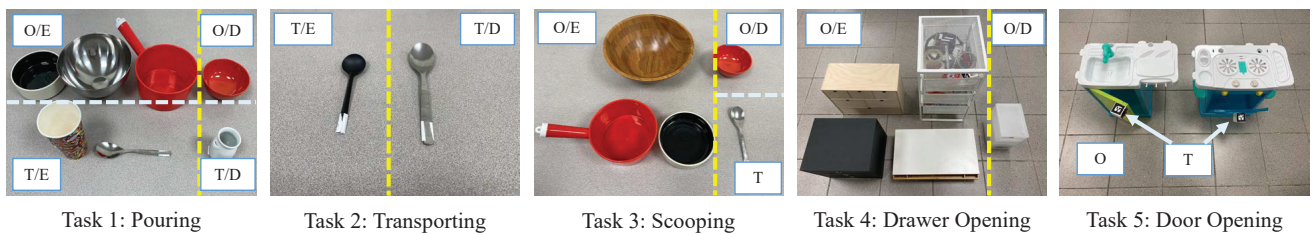}
\caption{Objects of interest and tools used for each task. $O$ and $T$ stand for ``object of interest'' and ``tool'', respectively; $E$ and $D$ stand for ``robot execution'' and ``demonstration'', respectively. The format $X/Y$ means ``$X$ used in the $Y$ process''. And the ones without $E$ or $D$ are used in both processes.}
\label{fig:experiments:object_tool}
\end{figure*}

\begin{figure*}
\centering
\includegraphics[scale=0.8, trim = 160 20 160 20, clip]{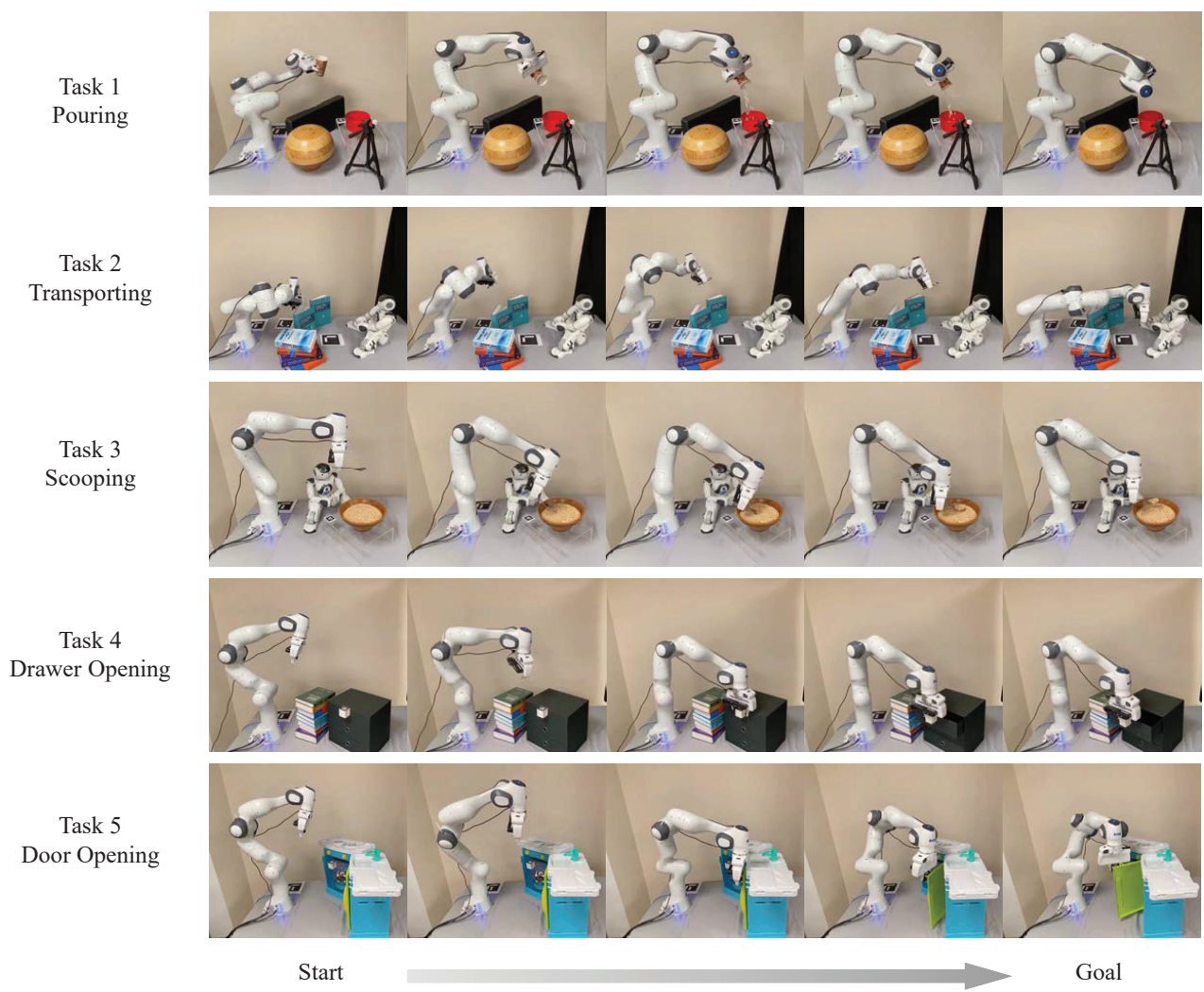}
\caption{Experiment results using the proposed robotic system including PRIMP, Workspace-STOMP and robot imagination.}
\label{fig:experiments:results}
\end{figure*}

The objects and tools used for each task are shown in Fig. \ref{fig:experiments:object_tool}. For the demonstrations and robot executions, different objects and tools are used. The experimental results for the tasks in Tab. \ref{tab:benchmark:tasks} are demonstrated in Fig. \ref{fig:experiments:results}. One example is shown for each task.

\section{Discussion and Future Work} \label{sec:discussion}
This section discusses the proposed PRIMP method, Workspace-STOMP planner and the robotic system for affordance learning from demonstration that transfers skills to novel scenarios.

\subsection{Features and Performance of PRIMP}
PRIMP learns the probability distribution and generalizes to novel situations in the full workspace of the robot manipulator, therefore it is easy to transfer between different robots. The learning process itself is robot-agnostic, but can be adapted to other robots with the conditioning by workspace density. Afterward, the learned trajectory distribution can be adjusted to the higher probability region in the robot workspace. When there is a rich set of demonstrations, the probability density gives more information for the whole motion where critical parts can be reflected. Even when there is only one demonstration, an initial covariance can be manually defined without affecting the generalization process for novel situations. Another feature of PRIMP is that the covariance is computed for the relative poses between two adjacent time steps. This is different from many existing probabilistic methods which compute covariance for absolute states. It is advantageous for the extrapolation case, where via poses are out of the range of demonstrations. When there is a change of the start pose, the whole trajectory will be shifted accordingly, but the internal connections between poses and the trajectory shape will remain invariant. The physical experiments demonstrate this feature since all new situations require the specific definition of a new start pose. When other via poses are far from the learned distribution, PRIMP can still perform well to both reach the target pose and maintain the trajectory shape, as shown in the qualitative examples. 

From the benchmark studies, PRIMP outperforms other probabilistic methods being compared in most cases, in terms of similarity metric for the whole trajectory as well as the distance to the desired via pose. The computational time for encoding and generalization can achieve the level of milliseconds, with an average of less than $50 \, ms$, which is more than 5 times faster than the compared counterparts. In terms of parameters, ProMP requires an explicit definition of the basis function and Orientation-KMP uses kernel tricks with parameters that affect performance significantly. However, the proposed PRIMP does not have any parameters to tune. Ours leads the performance ranking with a critical difference in LASA dataset, as shown in Fig. \ref{fig:benchmark:lfd:lasa:cd}, and outperforms in synthetic and real data for daily tasks. For the cases of $\SE(3)$ and $\PCG(3)$, PRIMP performs similarly, with $\PCG(3)$ being slightly better. This is partly because that when translation and orientation parts are separated, the mutual influence or error accumulation between different types of motions is mitigated. The suggestions for the use-cases of the two spaces are discussed as follows. If only the trajectory is considered without any kinematic constraint, $\PCG(3)$ is a conceptually more natural choice of representation. But if the motions are incrementally updated by system dynamics, such as for robots like cars or quadrotors, $\SE(3)$ can be a better choice.

\subsection{Performance of Workspace-STOMP}
The vanilla STOMP implementation without guidance runs faster with a high success rate, but always has larger errors in terms of similarity with the reference. In empty or sparse scenes, the proposed Workspace-STOMP algorithm runs as fast as the vanilla one with a similar success rate, but always has much smaller deviations with the reference trajectory. In more complex and narrow environments, the proposed planner is also competitive with Cartesian-STOMP, which only uses the mean trajectory, in maintaining the shape of the reference trajectory as well as the computational speed. The covariance information provided by PRIMP gives more flexibility in varying the samples and guides the optimization through critical regions. For example, when the covariance is large, the trajectory can deviate more from the reference and focuses more on collision avoidance. And in a certain region when the covariance is small, the planner will maintain the shape much closer to the reference. This cannot be fulfilled when only the mean trajectory is considered in the cost.

\subsection{The role of robot imagination}
In this work, the pouring and scooping tasks require robot imagination to generate key pose for PRIMP. When combining with the functionality of objects like bowls, spoons or cups, the definition of these via poses are natural and explainable. By hard-coding and simulating the pouring and scooping processes, the critical functional pose is well-defined to assist the whole framework. Physical experiments demonstrate the feasibility of the imagined key poses, with the guidance of which the proposed PRIMP method and Workspace-STOMP planner are able to successfully transfer the human skills to the robot for the same task in different unseen scenarios.

\subsection{Limitations and future work}
The proposed methods still have limitations. For example, the imagination is still pre-programmed, which might not truly reflect human behaviors. The mean trajectory learned from PRIMP is sometimes not smooth enough, which requires STOMP to optimize. Also, when there is an anomaly in the demonstration set, PRIMP might have undesirable performance. To potentially resolve these limitations, future work includes fusing demonstrations into the robot imagination module; adding velocity and/or acceleration into the state vector; possibly adding force information in the probabilistic model; and investigating ways to detect and filter data outliers before encoding demonstrations.

\section{Conclusion} \label{sec:conclusion}
This article presents \textit{PRobabilistically-Informed Motion Primitives (PRIMP)}, a learning-from-demonstration method that computes the probability distribution in the robot workspace. It only requires a few or even a single demonstration, and is able to adapt to new via poses (including start, goal and any point in between), a change of viewing frame and robot-specific workspace density. The learned trajectory distribution is then used to guide STOMP motion planner to avoid novel obstacles, resulting in the \textit{Workspace-STOMP} planner. Benchmark studies show the superiority among different popular LfD methods and guided motion planners. The applicability is demonstrated experimentally in a novel robotic system with the study of object affordance.

\section*{Acknowledgement}
The authors would like to thank Dr. Seng-Beng Ho and Mr. Jikai Ye for useful discussions. This work was supported by NUS Startup grants A-0009059-02-00, A-0009059-03-00, CDE Board account E-465-00-0009-01, National Research Foundation, Singapore, under its Medium Sized Centre Programme - Centre for Advanced Robotics Technology Innovation (CARTIN), sub award A-0009428-08-00 and AME Programmatic Fund Project MARIO A-0008449-01-00. The ideas expressed in this paper are solely those of the authors.

\section*{Appendix} \label{appendix}
\subsection{Operations for SE(3)} \label{appendix:se}
An element $g \in \SE(3)$ can be written in homogeneous matrix form, \ie $g \doteq \left( \begin{matrix}
R & {\bf t} \\
{\bf 0}^T & 1
\end{matrix} \right)$, where $R \in \SO(3)$ and ${\bf t} \in \IR^3$ are the rotation and translation parts, respectively. The exponential and logarithm mapping are simply the matrix exponential and logarithm. The adjoint operator for $\SE(3)$ is defined as 
\begin{equation}
Ad(g) \doteq \left( \begin{matrix}
R & \mathbb{O}_{3 \times 3} \\
TR & R
\end{matrix} \right) \in \IR^{6 \times 6} \,,
\label{eq:PRIMP:gora:adjoint_SE3}
\end{equation}
where $T \doteq \widehat{\bf t} = \left( \begin{matrix}
0 & -t_3 & t_2 \\
t_3 & 0 & -t_1 \\
-t_2 & t_1 & 0
\end{matrix} \right)$ is skew-symmetric.

\subsection{Operations for PCG(3)} \label{appendix:pcg}
An element in the Pose Change Group (PCG) \cite{chirikjian2018pose} is represented as a rotation and translation pair, \ie $g \doteq (R, {\bf t}) \in \PCG(3) = \SO(3) \times \IR^3$. The exponential mapping for $\PCG(3)$ is $\exp(\xxi) = \left( \exp(\widehat{\oomega}), {\bf t} \right)$, where $\xxi \doteq [\oomega^T, {\bf t}^T]^T \in \IR^6$ and $\widehat{\oomega} = \log(R) \in \mathfrak{so}(3)$. The logarithm mapping for $\PCG(3)$ is $\log(g) = \left( \log(R), {\bf t} \right) \in \mathfrak{pcg}(3)$ and $\log^{\vee}(g) = [\left( \log^{\vee}(R) \right)^T, {\bf t}^T]^T = \xxi \in \IR^6$. The relative pose between $g_1$ and $g_2$ becomes $\Delta_{1,2} \doteq (R_1^T R_{2}, {\bf t}_{2} - {\bf t}_1)$. The adjoint operator of $\PCG(3)$ can be computed as 
\begin{equation}
Ad(g) \doteq \left( \begin{matrix}
R & \mathbb{O}_{3 \times 3} \\
\mathbb{O}_{3 \times 3} & \II_3
\end{matrix} \right) \in \IR^{6 \times 6} \,.
\end{equation}

\subsection{Derivation of joint probability that encodes the demonstrations} \label{appendix:joint_pdf}
Given the absolute mean trajectory $\{\mu_0, \mu_1, ..., \mu_n\}$ and relative covariance $\{\Sigma_{0,1}, \Sigma_{1,2}, ..., \Sigma_{n-1, n}\}$, the joint probability can be computed as follows. Assuming the variation of $i^{\rm th}$ pose only depends on its two neighboring poses and the start pose is fixed, then
\begin{equation}
\rho(g_1, g_2,...,g_{n}) = \prod_{i=0}^{n-1} \rho(g_{i+1} | g_i) \,.
\label{eq:app:joint_pdf:condition_pdf}
\end{equation}

If all the poses are subject to Gaussian distribution with small deviations, explicit results can be shown as follows. Firstly, 
\begin{equation}
\begin{aligned}
& \rho(g_{i+1} | g_i) \\
&\propto \exp \left(-\frac{1}{2} \left\| \log^{\vee}[ (\mu_i^{-1} \circ \mu_{i+1})^{-1} \circ (g_i^{-1} \circ g_{i+1}) ] \right\|^2_{\Sigma_{i,i+1}} \right) \,.
\end{aligned}
\label{eq:app:joint_pdf:condition_pdf_gaussian}
\end{equation}
Defining free variables that measure the deviation of the true poses to their mean as $g_i = \mu_i \circ \exp(\widehat\xx_i)$. Then, the matrix logarithm term can be approximated as
\begin{equation}
\begin{aligned}
& \log^{\vee}\left[ (\mu_i^{-1} \circ \mu_{i+1})^{-1} \circ \left( (\mu_i \circ \exp(\widehat\xx_i))^{-1} \circ (\mu_{i+1} \circ \exp(\widehat\xx_{i+1})) \right) \right] \\
=& \,\, \log^{\vee}\left[ (\mu_i^{-1} \circ \mu_{i+1})^{-1} \circ \exp^{-1}(\widehat\xx_i) \circ (\mu^{-1} \circ \mu_{i+1}) \circ \exp(\widehat\xx_{i+1}) \right] \\
\approx& \,\, \xx_{i+1} - Ad_{i,i+1} \xx_i \,,
\end{aligned}
\label{eq:app:joint_pdf:condition_pdf_gaussian:log_approx}
\end{equation}
where $Ad_{i,i+1} \doteq Ad(\mu_i^{-1} \circ \mu_{i+1})$. Then, substituting Eq. \eqref{eq:app:joint_pdf:condition_pdf_gaussian:log_approx} into Eq. \eqref{eq:app:joint_pdf:condition_pdf_gaussian} and defining a joint variable $\xx_{i,i+1} \doteq [\xx_i^T, \xx_{i+1}^T]^T$ gives
\begin{equation}
\rho(g_{i+1} | g_i) \propto \exp \left(-\frac{1}{2} \left\| \xx_{i,i+1} \right\|^2_{\Sigma'_{i,i+1}} \right) \,,
\label{eq:app:joint_pdf:condition_pdf_gaussian_approx}
\end{equation}
where 
\begin{equation}
\Sigma'^{-1}_{i,i+1} = \left(\begin{matrix}
Ad^{-1}_{i,i+1} \Sigma^{-1}_{i,i+1} Ad^{-1}_{i,i+1} & -Ad^{-T}_{i,i+1} \Sigma^{-1}_{i,i+1} \\
-\Sigma^{-1}_{i,i+1} Ad^{-T}_{i,i+1} & \Sigma^{-1}_{i,i+1}
\end{matrix}\right) \in \IR^{12 \times 12} \,.
\end{equation}

Finally, substituting Eq. \eqref{eq:app:joint_pdf:condition_pdf_gaussian_approx} into Eq. \eqref{eq:app:joint_pdf:condition_pdf} gives
\begin{equation}
\begin{aligned}
\rho(g_1, g_2,...,g_{n}) &\propto \exp \left(-\frac{1}{2} \sum_{i=1}^{n-1} \left\| \xx_{i,i+1} \right\|^2_{\Sigma^{' \, -1}_{i,i+1}} \right) \\
&= \exp \left(-\frac{1}{2} \left\| \xx_{1,...,n} \right\|^2_{\Sigma^{' \, -1}_{1,...,n}} \right) \,,
\end{aligned}
\label{eq:app:joint_pdf:conditional_pdf_gaussian_full}
\end{equation}
where 
$$
\xx_{1,...,n} \doteq [\xx_1^T,..., \xx_i^T,..., \xx_{n}^T]^T \,,
$$
\begin{strip}
\begin{equation}
\Sigma'^{-1}_{1,...,n} = \left( \begin{matrix}
\Sigma^{-1}_{0,1} + \widetilde{\Sigma}_{1,2} & -Ad^{-T}_{1,2} \Sigma^{-1}_{1,2} & 0 & 0 & ... & 0 \\
-\Sigma^{-1}_{1,2} Ad^{-T}_{1,2} & \Sigma^{-1}_{1,2} + \widetilde{\Sigma}_{2,3} & -Ad^{-T}_{2,3} \Sigma^{-1}_{2,3} & 0 & ... & 0 \\
0 & -\Sigma^{-1}_{2,3} Ad^{-T}_{2,3} & ... & ... & ... & ... \\
0 & 0 & ... & ... & ... & 0 \\
... & ... & ... & ... & \Sigma^{-1}_{n-2,n-1} + \widetilde{\Sigma}_{n-1,n} & -Ad^{-T}_{n-1,n} \Sigma^{-1}_{n-1,n} \\
0 & ... & ... & 0 & -\Sigma^{-1}_{n-1,n} Ad^{-1}_{n-1,n} & \Sigma^{-1}_{n-1,n}
\end{matrix} \right) \,,
\end{equation}
\end{strip}
and $\widetilde{\Sigma}_{i,i+1} = Ad_{i,i+1} \Sigma_{i,i+1} Ad^{T}_{i,i+1}$.

\subsection{Derivation of equivariant property of the joint distribution of the whole trajectory} \label{appendix:equivariance}
The conditional probability $\rho(g_{i+1} | g_i)$ under the change of view is derived as follows. For Gaussian distribution,
\begin{equation}
\begin{aligned}
& \rho(g_{i+1} | g_i) \propto \\
& \exp \left( -\frac{1}{2} \left\| \log^{\vee} \left( (\mu^{-1}_i \mu_{i+1})^{-1} (g^{-1}_i g_{i+1}) \right) \right\|^2_{\Sigma^{-1}_{i,i+1}} \right) \,.
\end{aligned}
\label{eq:app:equivariance:conditional_pdf}
\end{equation}
When $h \in \SE(3)$ is the transformation from frame $O$ to $A$, for any $g \in \SE(3)$ that is viewed in $O$, $g^{o} = h^{-1} g h$ is the conjugated element that is viewed in frame $A$. Then, by direct computations, the matrix logarithm term  within the squared weighted norm is derived as
\begin{equation}
\begin{aligned}
& \log^{\vee} \left( (\mu^{-1}_i \mu_{i+1})^{-1} (g^{-1}_i g_{i+1}) \right) \\
=& \log^{\vee} \left( h (\mu^{o \, -1}_i \mu^{o}_{i+1})^{-1} g^{o \, -1}_i g^{o}_{i+1} h^{-1} \right) \\
=& \log^{\vee} \left( h (\mu^{o \, -1}_i \mu^{o}_{i+1})^{-1} \exp(-\xx^{o}_i) (\mu^{o \, -1}_i \mu^{o}_{i+1}) \exp(\xx^{o}_{i+1}) h^{-1} \right) \\
=& Ad(h) \log^{\vee} \left( (\mu^{o \, -1}_i \mu^{o}_{i+1})^{-1} \exp(-\xx^{o}_i) (\mu^{o \, -1}_i \mu^{o}_{i+1}) \exp(\xx^{o}_{i+1}) \right) \,.
\end{aligned}
\end{equation}
With the assumption of small deviations, \ie $\| \xx^{o}_i \|_2 \ll 1$, Eq. \eqref{eq:app:equivariance:conditional_pdf} can be approximated as
\begin{equation}
\begin{aligned}
& \rho(g_{i+1} | g_i) \propto \\
& \exp \left( -\frac{1}{2} \left\| Ad(h) \left( \xx^{o}_{i+1} - Ad^{o}_{i,i+1} \xx^{o}_i \right) \right\|^2_{\Sigma^{-1}_{i,i+1}} \right) \,,
\end{aligned}
\end{equation}
which is equivalent to
\begin{equation}
\rho(g_{i+1} | g_i) \propto \exp \left( -\frac{1}{2} \left\| \xx^{o}_{i+1} - Ad^{o}_{i,i+1} \xx^{o}_i \right\|^2_{ \Sigma^{o \, -1}_{i,i+1}} \right) \,,
\end{equation}
where,
$$
\Sigma^{o}_{i,i+1} \doteq Ad^{-1}(h) \Sigma_{i,i+1} Ad^{-T}(h) \,.
$$

\bibliographystyle{IEEEtran}
\bibliography{main}

\end{document}